\documentclass[letterpaper, 10 pt, conference]{ieeeconf}  

\IEEEoverridecommandlockouts                              

\usepackage{amsmath} 
\usepackage{amssymb}  
\usepackage{tcolorbox}
\usepackage{makecell}
\usepackage{bbding}
\usepackage{booktabs}
\usepackage{graphicx}       
\usepackage[table]{xcolor}  
\usepackage{multirow}
\usepackage{float}
\usepackage{graphicx}
\usepackage{subcaption} 
\usepackage{caption}    
\usepackage{url}
\usepackage[misc]{ifsym}
\usepackage{fix-cm}
\usepackage{hyperref}

\title{\LARGE \bf
VO-DP: Semantic-Geometric Adaptive \textcolor{orange}{D}iffusion \textcolor{orange}{P}olicy for \\ \textcolor{orange}{V}ision-\textcolor{orange}{O}nly Robotic Manipulation
}

\author{
Zehao Ni$^{1,2,5,6}$\textsuperscript{*},
Yonghao He$^{1}$\textsuperscript{*,$\dagger$},
Lingfeng Qian$^{1}$\textsuperscript{*},
Jilei Mao$^{1}$,
Fa Fu$^{1}$,
Wei Sui$^{1}$, \\
Hu Su$^{4}$,
Junran Peng$^{3}$\textsuperscript{\Letter},
Zhipeng Wang$^{2,5,6}$,
Bin He$^{2,5,6}$\textsuperscript{\Letter}
\\
\\
$^{1}$ \textsc{D-Robotics} \\
$^{2}$ National Key Laboratory of Autonomous Intelligent Unmanned Systems \\
$^{3}$ University of Science and Technology Beijing \\
$^{4}$ State Key Laboratory of Multimodal Artificial Intelligence System (MAIS) \\
\hspace*{1.1em}Institute of Automation of Chinese Academy of Sciences \\
$^{5}$ Frontiers Science Center for Intelligent Autonomous Systems \\
$^{6}$ Shanghai Institute of Intelligent Science and Technology, Tongji University \\
\\
\textsuperscript{*}Equal contribution \quad 
\textsuperscript{$\dagger$}Project lead \quad 
\textsuperscript{\Letter}Corresponding author
\\
\href{https://d-robotics-ai-lab.github.io/vodp/}{d-robotics-ai-lab.github.io/vodp}
}

\begin{document}

\maketitle
\thispagestyle{empty}
\pagestyle{empty}

\begin{abstract}

In the context of imitation learning, visuomotor-based diffusion policy learning is one of the main directions in robotic manipulation.
Most of these approaches rely on point clouds as observation inputs and construct scene representations through point clouds feature learning, which enables them to achieve remarkable accuracy. 
However, the existing literature lacks an in-depth exploration of vision-only solutions that have significant potential.
In this paper, we propose a \textbf{V}ision-\textbf{O}nly and single-view \textbf{D}iffusion \textbf{P}olicy learning method (\textbf{VO-DP}) that leverages pretrained visual foundation models to achieve effective fusion of semantic and geometric features. 
We utilize intermediate features from VGGT incorporating semantic features from DINOv2 and geometric features from Alternating Attention blocks. Features are fused via cross-attention and spatially compressed with a CNN to form the input to the policy head.
Extensive experiments demonstrate that VO-DP not only outperforms the vision-only baseline DP significantly but also exhibits distinct performance trends against the point cloud-based method DP3:
in simulation tasks, VO-DP achieves an average success rate of 64.6\%—on par with DP3 64.0\% and far higher than DP 34.8\%, while in real-world tasks, it reaches 87.9\%, outperforming both DP3 67.5\% and DP 11.2\% by a notable margin.
Further robustness evaluations confirm that VO-DP remains highly stable under varying conditions including color, size, background, and lighting.
Lastly, we open-source DRRM (D-Robotics Robotic Manipulation), a training library for robotic manipulation. Built on Accelerate, this library supports multi-machine and multi-GPU parallel training, as well as mixed precision training (e.g., bf16, fp16). It is compatible with visuomotor policies such as DP and DP3, and also supports the RoboTwin simulator. VO-DP is integrated into DRRM. We refer to the \href{https://d-robotics-ai-lab.github.io/vodp/}{project page} for the code and videos. 

\end{abstract}


\section{INTRODUCTION}
Visuomotor policy learning has emerged as an important paradigm in robotic manipulation, leveraging visual observations to guide the generation of action sequences in an end-to-end manner. 
Current mainstream visuomotor methods can be broadly categorized into non-vision-only approaches and vision-only approaches.
\begin{figure}[!htbp]
    \centering
    \includegraphics[width=\linewidth]{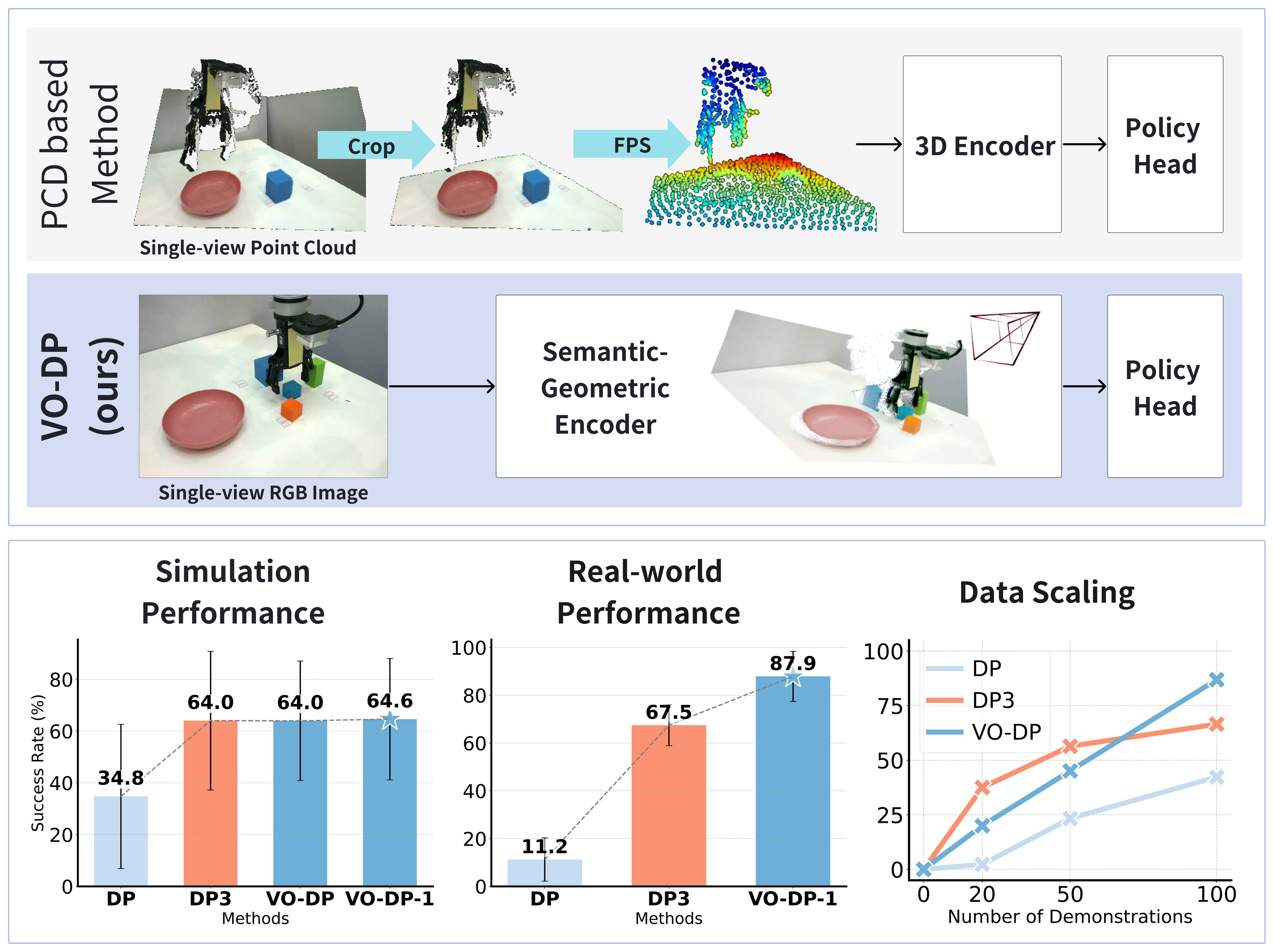}
    \caption{VO-DP is a vision-only method for visuomotor robotic manipulation: it takes single-view RGB images as input, uses large vision models to extract semantic and geometric features from observations, and provides high-quality conditional inputs for the policy head. Experiments show it matches point cloud-based DP3’s accuracy in simulation, outperforms it significantly in real-world tasks, and notably boosts vision-only method accuracy.}
    \label{fig:vo}
\end{figure}
Vision-only approaches rely on RGB image inputs to achieve joint optimization of perception and action. 
Such methods depend on implicit 3D scene understanding and closely align with biological perception-action systems.
In contrast, non-vision-only approaches rely on explicit 3D representations, such as point clouds (e.g., DP3 \cite{ze20243d}, 3D Diffuser Actor \cite{ke20243d}) or RGB-D images (e.g., SEM \cite{lin2025sem}, H$^3$DP \cite{lu2025h}) as inputs to decouple the 3D modeling process. 
Benefiting from precise low-dimensional 3D representations, these methods have significantly improved the accuracy of robotic manipulation.
However, non-vision-only methods heavily depend on high-cost hardware and exhibit limitations in complex scenes. 
First, acquiring RGB-D or point clouds requires expensive sensors such as depth cameras or LiDAR, and model performance is constrained by sensor accuracy. 
In comparison, RGB cameras provide significantly lower costs and higher practicality: their hardware costs are reducible by orders of magnitude, and further, system complexity arising from multi-sensor calibration is avoided.
Second, experimental results demonstrate that sparse 3D inputs are inadequate for semantic-intensive tasks and complex scenarios, which in turn leads to performance degradation.

We argue that the academic community lacks in-depth research and exploration under vision-only settings, particularly regarding how to learn effective representations for robotic manipulation when relying solely on RGB images. 
Currently, vision-only methods have not yet demonstrated performance superior to that of point cloud-based methods in robotic manipulation.
This is largely attributed to underdeveloped representation learning modules in existing methods. 
To further unlock the potential of vision-only approaches, we propose VO-DP, a method that integrates and compresses both semantic and geometric features extracted from single-view image as input to a downstream policy head.
Specifically, we leverage intermediate-layer features from the pretrained 3D reconstruction model VGGT \cite{wang2025vggt}, including semantics-aware features from DINOv2 \cite{oquab2023dinov2} and geometry-aware features from the Alternating Attention network. 
We then design a cross-attention-based fusion module to adaptively inject semantics-aware features into geometry-aware features according to task-specific information preferences. 
Finally, we introduce a spatial feature compression module, which is based on CNN, to distill essential representations from the scene.

In summary, our contributions are four-fold:
\begin{itemize}
    \item We demonstrate that vision-only visuomotor policies hold substantial performance potential in robotic manipulation, even achieving an accuracy level on par with point cloud-based methods.
    \item We propose VO-DP, a novel vision-only, single-view representation learning visuomotor method for robotic manipulation that adaptively fuses semantic and geometric information.
    \item We conduct a detailed evaluation and analysis of the VO-DP method on the RoboTwin 1.0 \cite{mu2024robotwin} simulation benchmark and real-world tasks, and it achieves state-of-the-art performance in both simulation and real-world experiments.
    \item We open-source a general training framework for robotic manipulation. Built on Accelerate\cite{gugger2022accelerate}, it supports multi-node multi-GPU training and multi-GPU evaluation with the RoboTwin simulator, offers mixed-precision training (bf16/fp16), and maintains compatibility with visuomotor methods such as DP and DP3.
\end{itemize}

\section{RELATED WORK}

\subsection{Vision-Only Methods}
Vision-only methods refer to those approaches that take RGB images as observation inputs.
Some typical examples of such methods, including DP\cite{chi2023diffusion},ACT\cite{zhao2023learning} and others \cite{brohan2022rt,chi2023diffusion,jang2022bc,zhao2023learning,shafiullah2022behavior,mao2025omnid}, have demonstrated exceptional performance in robotic manipulation tasks. 
However, it has been observed that while these methods deliver reasonably good performance in in-distribution scenarios, they show notable sensitivity to environmental variations during real-world deployment. 
Even minor changes in background, camera pose, or lighting conditions can trigger severe degradation in model performance\cite{xie2024decomposing}.
Within image-based approaches, the impact of representation learning has received relatively little attention in existing studies. While DP\cite{chi2023diffusion}, one of our baseline methods, does explore how different backbones influence success rates, its analysis remains confined to conventional image backbones such as ResNet\cite{he2016deep} and ViT\cite{dosovitskiy2020image}, with no extension to specialized or advanced representation learning architectures.
We contend that, for robotic manipulation tasks, learning appropriate representations from RGB image-based inputs plays a pivotal role in enhancing the robustness and generalization capabilities of vision-only models. 

Recent years have witnessed the rapid development of visual foundation models\cite{hong20233d,fu2024scene,zhu2024llava}.
In particular, visual models with spatial perception capabilities, such as VGGT\cite{wang2025vggt}, can directly extract geometric information from RGB images, thereby providing rich feature options for vision-only methods.
OV-DP leverages the semantic and geometric features provided by VGGT, fully unlocking the potential of vision-only methods without additional preprocessing that 3D-based methods frequently rely on.

\subsection{Non-Vision-Only Methods}
Non-vision-only methods refer to those that take 3D signals, such as depth information and point clouds, as observation inputs.
PerAct\cite{shridhar2023perceiver} voxelizes point clouds data into tokens, which are then fused with text tokens in a transformer architecture. 
ACT3D\cite{gervet2023act3d} utilizes the CLIP\cite{radford2021learning} model to extract image features and further aggregates them with depth information to enhance model performance. 
3D Diffuser Actor\cite{ke20243d} converts 2D image features into 3D tokens using a depth map and employs a denoising network to generate trajectories. 
RVT\cite{goyal2023rvt} projects RGB-D data on three orthogonal planes to form virtual images, which are subsequently used for action prediction. 
RVT-2\cite{goyal2024rvt} optimizes the prediction head by leveraging heatmaps for trajectory generation, enabling more precise manipulation. 
DP3\cite{ze20243d} takes single-camera point clouds as input and generates actions through a sequence of preprocessing steps—including point clouds filtering, clustering, feature extraction, and denoising. Notably, this point clouds preprocessing pipeline is not only complex but also relies on high-precision RGB-D cameras.
To underscore the superiority of our proposed method, we thus select DP3 as one of our baseline methods.

Non-vision-only methods exhibit strong performance in both success rate and few-shot learning. However, they not only depend on intricate point clouds preprocessing pipelines but also often require accurate camera extrinsic calibration for reprojection, two factors that complicate real-world deployment and place high demands on sensor consistency and environmental stability. 
By contrast, our approach relies solely on image input, yet delivers accuracy on par with these non-vision-only methods.
\begin{figure*}[!htbp]
    \centering
    \includegraphics[width=\textwidth]{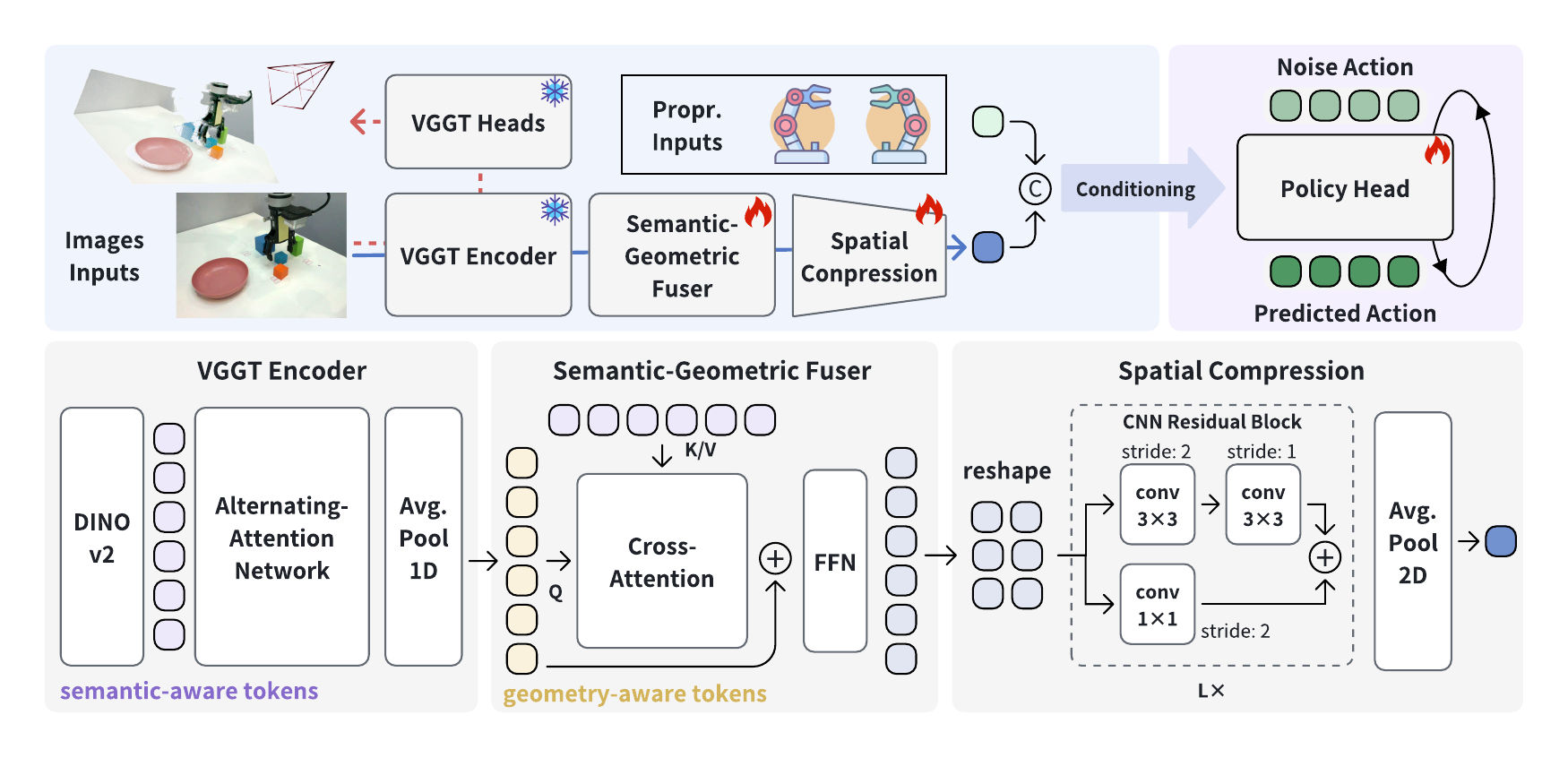}
    \vspace{-8mm}
    \caption{Overall architecture of VO-DP. VO-DP has four core modules: 1) VGGT Encoder extracts semantic features from patchified images via DINOv2 and generates geometric features through its AA network; 2) Semantic-Geometric Fuser fuses per-frame geometric and semantic features using residual cross-attention and an FFN; 3) Spatial Compression module reshapes fused features, downsamples them with a lightweight ResNet, and concatenates the compressed spatial features with proprioceptive observations to form compact scenario representations; 4) Vision-Only Conditioned Action Generation module employs a DDPM-based policy head to generate actions using the scenario representations.}
    \vspace{-3mm}
    \label{fig:VODP}
\end{figure*}

\section{METHOD}
We define the vision-only imitation learning task as follows: given a small set of expert demonstrations containing both video and robotic action trajectories, learn a visuomotor policy $\pi$ that maps observations $O_t \in \mathcal{O}$ at time step $t$ to actions $A_t \in \mathcal{A}$. 
The observation $O_t$, as shown in (\ref{eq:obs}), consists of an RGB image history of size $T$ and a sequence of  $T$ states of $J$ joints, and the output $A_t \in \mathbb{R}^{N \times J}$ represents a predicted action trajectory of length $N$, 
\begin{equation}
    \label{eq:obs}
    O_t = \{ I_t \in \mathbb{R}^{T \times H \times W \times 3}, S_t \in \mathbb{R}^{T \times J} \}.
\end{equation}
Notably, our method uses only single-view RGB images. The overall architecture is illustrated in \ref{fig:VODP}, which consists of four sub-modules: 
a pretrained visual encoder \ref{subsec:encoder} incorporating geometric priors for encoding observation images into semantic and geometric features; 
a semantic-geometric feature fusion module \ref{subsec:fuser} for adaptive modality selection tailored to specific tasks; 
a scenario representation compressor \ref{subsec:compact} for distilling key information of scene; 
and a policy head \ref{subsec:head} that predicts action chunks conditioned on the scenario features.

\subsection{Geometry Prior-based Visual Encoder}
\label{subsec:encoder}
Benefiting from its concise architecture and robust generalization capability, VGGT \cite{wang2025vggt} is employed as the visual encoder in our method.
Pretrained on a variety of 3D reconstruction tasks, VGGT can extract essential geometric features directly from one or a few input images and predict comprehensive 3D attributes of a scene — including camera parameters, point clouds, depth maps, and 3D point tracks. 
Specifically, VGGT is implemented as a large transformer \cite{vaswani2017attention}, where each input image is first patchified into a set of tokens via DINOv2 \cite{oquab2023dinov2}. 
The combined image tokens from all input frames are then processed through an Alternating-Attention (AA) network consisting of 24 AA blocks. 
Each AA block is designed with a frame-wise self-attention layer followed by a global self-attention layer. 
The features output by the AA network are further fed into prediction heads to estimate 3D attributes.


Therefore, we posit that the features processed by the AA network encapsulate rich 3D geometric information, which can enhance the precision of vision-only visuomotor policies. 
In our implementation, the image history $I_t$ is patchified via DINOv2 into a set of $T \times P$ tokens, denoted as $\mathbf{h^{sem}_t} \in \mathbb{R}^{T \times P \times C}$, where $C$ is the feature dimension. $\mathbf{h^{sem}_t}$ serves as semantic features of the observation. 
These tokens are then fed into the pretrained AA network, and the output from the 24-$th$ AA block is adopted. 
Notably, within each AA block, VGGT concatenates features derived from both the frame-wise and global self-attention layers prior to feeding them into the prediction heads, thus enabling the integration of local and global information.
Accordingly, we use these concatenated features as the geometric feature representation $\mathbf{h_t^{geo}} \in \mathbb{R}^{T \times P \times 2C}$.

\subsection{Semantic-Geometric Feature Fuser}
\label{subsec:fuser}
To effectively leverage both semantic features and geometric information, we fuse the per-frame features $\mathbf{g} = \mathbf{h_t^{geo}[i]} \in \mathbb{R}^{P \times 2C}$ and $\mathbf{s}  = \mathbf{h^{sem}_t[i]} \in \mathbb{R}^{P \times C}$ using residual cross-attention, where $i$ is the frame index, $\mathbf{g}$ serves as the query and $ \mathbf{s}  $ as the key-value pair, as follows:
\begin{equation}
    \label{eq:crossattn}
    \begin{aligned}
    \mathbf{h^\prime} &= \rm{AvgPool} \mit(\mathbf{g}), \\
    \mathbf{h^{\prime\prime}} &= \mathbf{h^\prime} + \rm{CrossAttn} \mit{(\mathbf{h^\prime} \mathbf{W_Q},\mathbf{s}  \mathbf{W_K},\mathbf{s}  \mathbf{W_V})}, \\
    \end{aligned}
\end{equation}
where $\rm{AvgPool}(\cdot)$ performs feature compression using 1D average pooling with a kernel size of 2 and stride of 2 along the feature dimension, $\mathbf{W_Q} \in \mathbb{R}^{C \times C}$, $\mathbf{W_K} \in \mathbb{R}^{C \times C}$, and $\mathbf{W_V} \in \mathbb{R}^{C \times C}$ are projection matrices.
The features after cross-attention are further projected through a Feed-Forward Network (FFN) layer:
\begin{equation}
    \label{eq:ffn}
    \mathbf{h^{sg}_t[i]} = \mathbf{h^{\prime\prime}} + \rm{FFN}\mit(\mathbf{h^{\prime\prime}}),
\end{equation}
where $\mathbf{h^{sg}_t[i]}$ is the fused features with semantic and geometric information.

\subsection{Scenario Representation Compression}
\label{subsec:compact}
We then encode all observation tokens into compact scenario representations with a lightweight ResNet \cite{he2016deep}, as shown in Fig. \ref{fig:VODP}. 
The layout of $\mathbf{h^{sg}_t}$ is first reshaped to $\mathbb{R}^{T \times C \times H_P \times W_P}$, where $H_P$, $W_P$ denote the height and width of patch grid. 
We then apply three basic residual blocks, each with a kernel size of 3 and a stride of 2, to downsample the feature maps. 
An adaptive 2D average pooling layer compresses the remaining patches into a spatial feature $\mathbf{h^{sp}_t}$, which is projected into a low-dimensional space $C'$ and concatenated with the proprioceptive observation $\mathbf{S_t}$, yielding the scene representation $\mathbf{h^{sc}_t} \in \mathbb{R}^{T \times (C'+J)}$,
\begin{equation}
    \label{eq:scenario}
    \mathbf{h^{sc}_t} = \left[ \rm{MLP} \mit(\mathbf{h^{sp}_t)}, \mathbf{S_t}  \right].
\end{equation}

\subsection{Vision-Only Conditioned Action Generation}
\label{subsec:head}
For the policy head, we follow the original DP implementation \cite{chi2023diffusion}, training it with a vision-only conditioned Denoising Diffusion Probabilistic Model (DDPM) \cite{ho2020denoising} to approximates the conditional distribution $p(A_t | O_t)$ introduced in \cite{janner2022planning}. The denoising process follows:
\begin{equation}
    \label{eq:process}
    A_t^{k-1} = \alpha (A^k_t - \gamma \varepsilon_\theta(h^{sc}_t ,A^k_t, k) + \mathcal{N}(0,\sigma^2I))
\end{equation}
where $\gamma$ is the learning rate, $\alpha$ and $\sigma$ are scalar coefficients predefined by a noise scheduler, and $\varepsilon_\theta$ is the noise prediction network with parameters $\theta$ which predicts the noise of the trajectory $A^k_t$ conditioned on the scene feature $h^{sc}_t$. We train it with MSE loss as follows:
\begin{equation}
    \label{eq:loss}
    \mathcal{L}(\theta) = \rm{MSE}(\varepsilon^k, \varepsilon_\theta(h^{sc}_t ,A^k_t, k)).
\end{equation}
\section{SIMULATION EXPERIMENTS}
\begin{figure}[!thbp]
    \centering
    \includegraphics[width=\linewidth]{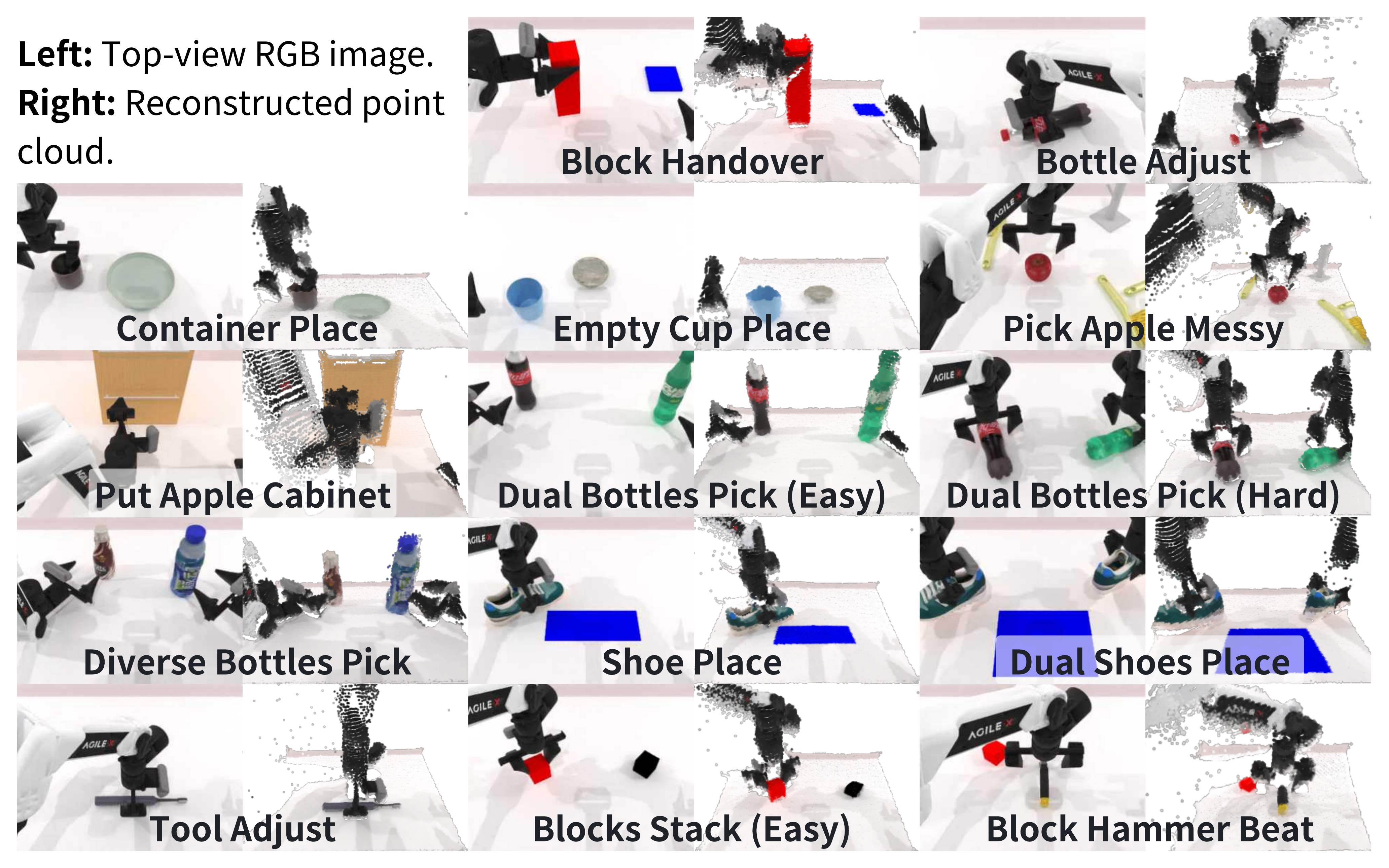}
    \caption{Simulation benchmark -- 14 bimanual manipulation tasks. Left: Top-view RGB image of the task. Right: Reconstructed point clouds via VGGT.}
    \vspace{-5mm}
    \label{fig:robotwin}
\end{figure}
\subsection{Experimental Setup}
\textbf{Benchmark.} Simulation experiments are conducted using the RoboTwin \cite{mu2024robotwin} benchmark, which is built upon the SAPIEN \footnote{SAPIEN: https://sapien-sim.github.io/docs/} simulator and comprises 14 bimanual manipulation tasks, as shown in Fig.~\ref{fig:robotwin}. 
RoboTwin serves as a challenging benchmark that requires manipulation policies to comprehend both semantic intent and geometric structure in visually complex environments.
RoboTwin employs an open-source Cobot Magic platform\footnote{Platform Introduction: https://global.agilex.ai/products/cobot-magic}, controlled via an action vector $J=14$ to operate a pair of 6-DoF simulated robotic arms equipped with grippers. 
Visual input consists of RGB-D images and point clouds captured by an Intel RealSense D-435 camera at a resolution of $240 \times 320$.

\textbf{Data Collection.} Training data are collected from 100 valid scenes initialized randomly starting from seed 0 for each task. 
During testing, evaluation is performed over 100 valid scenes initialized from seed 10000 per task, with each scene repeated three times. 
The success rate is determined based on satisfying target pose constraints upon task completion and maintaining collision-free execution throughout the trajectory.

\begin{table}[!htbp]
    \caption{Training Hyperparameter Settings.}
    \label{tab:hyperparams}
    \centering
    \footnotesize
    \setlength{\tabcolsep}{6pt}
    \belowrulesep=0pt
    \aboverulesep=0pt
    \resizebox{1.0\linewidth}{!}{
        \renewcommand{\arraystretch}{1.2}
        \begin{tabular}{l|c|l|c}
            \toprule
            \textbf{Hyperparam.} & \textbf{Value} & \textbf{Hyperparam.} & \textbf{Value} \\
            \midrule
            \cellcolor{yellow!15}batch size & \cellcolor{yellow!15}128  & \cellcolor{blue!15}adam beta1 & \cellcolor{blue!15}0.95\\
            \cellcolor{yellow!15}mixed\_precision & \cellcolor{yellow!15}bf16 & \cellcolor{blue!15}adam beta2 & \cellcolor{blue!15}0.99\\
            \cellcolor{red!15}learning rate & \cellcolor{red!15}1e-4 & \cellcolor{blue!15}adam\_weight\_decay & \cellcolor{blue!15}1e-6\\
            \cellcolor{red!15}lr scheduler & \cellcolor{red!15}cosine & \cellcolor{blue!15}adam\_epsilon & \cellcolor{blue!15}1e-8\\
            \cellcolor{red!15}lr\_warmup\_ratio & \cellcolor{red!15}0.05 & \cellcolor{green!15}ema: inv\_gamma & \cellcolor{green!15}1.0\\
             &  & \cellcolor{green!15}ema: power & \cellcolor{green!15}0.75\\
            \bottomrule
        \end{tabular}
    }
    \vspace{-3mm}
\end{table}

\textbf{Training Details.} The proposed method learns a mapping from an observation $O_t$ to an action trajectory $A_t$ of length $N=8$. To maintain consistency with DP and DP3, the observation adopts a history length of $T=3$ (denoted as VO-DP); additionally, the method is evaluated under an ablated setting with $T=1$ (denoted as VO-DP-1).
All models are trained for 300 epochs. 
During training, all samples are generated using the same random seed to ensure consistency. 
The models are trained on 8 NVIDIA A100 GPUs using the bfloat16 precision. Detailed hyperparameters are provided in Table \ref{tab:hyperparams}.

\begin{table*}[!htbp]
    \caption{RoboTwin Benchmark Results. DP: Diffusion policy\cite{chi2023diffusion}, DP3: 3D diffusion policy\cite{ze20243d}, VO-DP: $T=3$, VO-DP-1: $T=1$}
    \label{tab:main_result}
    \centering
    \footnotesize
    \setlength{\tabcolsep}{4pt}
    \belowrulesep=0pt
    \aboverulesep=0pt
    \resizebox{1.0\linewidth}{!}{
        \renewcommand{\arraystretch}{1.2}
        \begin{tabular}{c|c|c|c|c|c}
            \toprule
            \textbf{Method} & \textbf{Block Hammer Beat} & \textbf{Block Handover} & \textbf{Bottle Adjust} & \textbf{Container Place} & \textbf{Empty Cup Place} \\
            \midrule
            DP & 0.7±0.9 & 77.7±4.5 & 39.3±0.5 & 14.0±6.9 & 69.3±2.5 \\
            \rowcolor{red!15} DP3                 & 79.3±1.2          & \textbf{97.7±1.2} & \textbf{85.3±0.5} & \textbf{83.7±1.7} & \textbf{88.7±1.7} \\
            \rowcolor{cyan!15} \textbf{VO-DP}     & \textbf{85.0±1.4} & 89.7±0.5          & 63.3±1.2          & 43.0±3.7          & 82.0±2.2 \\
            \rowcolor{cyan!15} \textbf{VO-DP-1}   & 78.7±5.2          & \textbf{94.7±0.5} & 69.3±2.5          & 31.3±2.6          & 77.3±1.7 \\
            \midrule
            \textbf{Method} & \textbf{Pick Apple Messy} & \textbf{Put Apple Cabinet} & \textbf{Dual Bottles Pick (Easy)} & \textbf{Dual Bottle Pick (Hard)} & \textbf{Diverse Bottles Pick} \\
            \midrule
            DP & 31.0±0.8 & 63.6±1.9 & 73.7±1.2 & 63.3±0.5 & 7.3±1.2 \\
            \rowcolor{red!15} DP3 & 18.7±2.9 & 84.7±0.5 & 83.3±0.5 & 64.0±0.8 & \textbf{60.7±0.5} \\
            \rowcolor{cyan!15} \textbf{VO-DP}   & \textbf{80.0±0.8} & \textbf{94.3±2.3} & \textbf{88.3±0.9} & \textbf{67.3±3.3} & 32.3±3.3 \\
            \rowcolor{cyan!15} \textbf{VO-DP-1} & \textbf{81.7±0.9} & \textbf{98.0±0.8} & \textbf{86.3±0.5} & 60.3±1.2 & 31.3±1.7 \\
            \midrule
            \textbf{Method} & \textbf{Shoe Place} & \textbf{Dual Shoes Place} & \textbf{Tool Adjust} & \textbf{Blocks Stack (Easy)} & \cellcolor{yellow!15} \textbf{AVG. (↑)} \\
            \midrule
            DP & 19.3±1.2 & 4.7±0.5 & 20.0±2.9 & 3.7±1.2 & \cellcolor{yellow!15} 34.8 \\
            \rowcolor{red!15} DP3               & \textbf{56.3±1.7} & 13.7±1.7          & \textbf{58.3±0.5} & 22.0±2.2          & \textbf{64.0} \\
            \rowcolor{cyan!15} \textbf{VO-DP}   & 43.0±0.8          & \textbf{17.0±0.8} & \textbf{58.3±3.9} & \textbf{52.3±2.5} & \textbf{63.9} \\
            \rowcolor{cyan!15} \textbf{VO-DP-1} & 52.0±0.8          & \textbf{19.3±0.9} & 55.3±2.6          & \textbf{69.3±2.5} & \textbf{64.6} \\
            \bottomrule
        \end{tabular}
    }
\end{table*}
\subsection{Simulation Performance}
By comparing our method with DP (a traditional vision-only method) and DP3 (a native point cloud-based method), we validate the effectiveness of the pretrained fusion perception approach for vision-only robotic manipulation. 
Compared to DP, our method achieves a substantial performance improvement. 
Furthermore, compared to DP3, it achieves comparable manipulation accuracy at lower hardware costs, as shown in Table \ref{tab:main_result}.

\textbf{Compared to DP,} VO-DP and VO-DP-1 achieve dramatic performance improvements across all tasks, with particularly notable gains in several key scenarios. For instance, in the \textit{Pick Apple Messy} task, their success rates rise substantially from the baseline of 31.0\% to over 80.0\%. In more complex tasks, such as \textit{Block Hammer Beat} and \textit{Blocks Stack (Easy)}, performance also improves sharply: starting from very low baselines of 0.7\% and 3.7\%, VO-DP and VO-DP-1 reach success rates of 85.0\% and 69.3\%, respectively.

These results strongly demonstrate that by integrating pretrained visual foundation models and a feature fusion-based perception strategy, VO-DP-1 significantly enhances the understanding of complex scenes and objects as well as the precision of manipulation. This effectively overcomes the inherent limitations of traditional vision-only methods, specifically their inadequacies in perceptual accuracy and generalization.

\textbf{Compared to DP3,} which relies on raw 3D point cloud inputs, VO-DP only requires a lower-cost monocular camera, yet achieves comparable or even superior overall performance.
In terms of average success rate (AVG), VO-DP (63.9\%) achieves near-parity with DP3 (64.0\%), while its single-frame variant—VO-DP-1 (64.6\%), outperforms DP3.
Importantly, VO-DP achieves top or joint-top performance across multiple key tasks, such as 
\textit{Block Hammer Beat}, \textit{Put Apple Cabinet}, and \textit{Blocks Stack (Easy)}.
This indicates that VO-DP effectively bridges the performance gap with 3D perception-based methods, relying solely on image data, by leveraging advanced visual models.
In the \textit{Pick Apple Messy} task, our method boosts the success rate by 63.0\% relative to DP3, demonstrating the advantages of pretrained implicit spatial representations for perceiving complex scenes.

Although VO-DP slightly underperforms DP3, a raw point cloud-based baseline, in some structured tasks or those requiring precise geometric information (e.g., \textit{Diverse Bottles Pick}), it still maintains high performance with only RGB image input, thereby significantly lowering hardware sensor requirements.

\textbf{Comparison between 3 frame and 1 frame variants.} VO-DP and VO-DP-1 exhibit comparable overall performance, with each demonstrating distinct strengths across different scenarios. Given that VO-DP-1 achieves slightly superior performance to VO-DP, VO-DP-1 is selected as the method for subsequent real-world experiments.

\subsection{Ablation Study}
We select five tasks to conduct ablation studies: \textit{Pick Apple Messy} (PAM), \textit{Block Hammer Beat} (BHB), \textit{Dual Bottles Pick (Easy)} (DBPE), \textit{Put Apple Cabinet} (PAC), and \textit{Blocks Stack (Easy)} (BSE) from RoboTwin. 
From a semantic standpoint, these tasks cover a spectrum of robotic manipulation types, involving varied objects and diverse scenarios.
From a geometric perspective, these tasks necessitate the robot to resolve spatial relationships—underscoring the core challenges in spatial perception and motion coordination.

\begin{table}[t]
    \caption{Ablation study on different modality features.}
    \label{tab:modality}
    \centering
    \footnotesize
    \setlength{\tabcolsep}{6pt}
    \belowrulesep=0pt
    \aboverulesep=0pt
    \resizebox{0.9\linewidth}{!}{
        \renewcommand{\arraystretch}{1.2}
        \begin{tabular}{c|c|c|c}
            \toprule
            \textbf{Module} & \textbf{PAM} & \textbf{BHB} & \textbf{DBPE} \\
            \midrule
            w/o geo. & 44.3±0.9 & 59.3±0.5 & \textbf{95.3±0.9} \\
            w/o sem. & 38.7±1.7 & 60.7±4.9 & 81.3±0.5 \\
            VO-DP  & \textbf{80.0±0.8} & \textbf{85.0±1.4} & \textbf{88.3±0.9} \\
            \midrule
            \textbf{Module} & \textbf{PAC} & \textbf{BSE} & \cellcolor{yellow!15} \textbf{AVG. ($\uparrow$)} \\
            \midrule
            w/o geo. & \textbf{98.0±0.8} & \textbf{58.7±0.9} & \cellcolor{yellow!15} 71.12 \\
            w/o sem. & 93.7±2.0 & 45.3±2.5 & \cellcolor{yellow!15} 63.9 \\
            VO-DP  & \textbf{94.3±2.3} & \textbf{52.3±2.5} & \cellcolor{yellow!15} \textbf{80.0} \\
            \bottomrule
        \end{tabular}
    }
\end{table}
\textbf{Comparison of different modality features}. 
VO-DP leverages both semantic and geometric features to enable spatial understanding. 
To evaluate the contribution of each feature modality, we perform an ablation study on the fusion module, retaining only semantics-aware features (denoted as "w/o geo.") or geometry-aware features (denoted as "w/o sem."), with results summarized in Table \ref{tab:modality}.
Results show that the full VO-DP model achieves the best overall performance, with an average success rate evidently higher than that of either ablated variant, confirming the effectiveness of our multimodal feature fusion design.
Specifically, in tasks demanding strong semantic understanding (e.g., \textit{Pick Apple Messy}, \textit{Block Hammer Beat}), performance drops substantially when semantic features are removed (w/o sem.), underscoring the critical role of semantic priors in object recognition and task reasoning.
For structurally complex tasks (e.g., \textit{Dual Bottles Pick (Easy)}), the removal of geometric features (w/o geo.) leads to noticeable performance degradation, highlighting the importance of spatial structure for bimanual coordination.
Notably, certain tasks (e.g., \textit{Put Apple Cabinet}) achieve relatively high performance with only a single modality, indicating that perceptual demands vary across different tasks. Nevertheless, VO-DP consistently outperforms both ablated models in most scenarios, which demonstrates that our fusion mechanism robustly generalizes to diverse manipulation requirements.

\textbf{Comparison of Downsampling Strategies for Geometric Features. }
When integrating VGGT geometric features, we explore two downsampling strategies: average pooling and an MLP-based projection, to reduce the dimensionality of geometry-aware tokens to 1024 dimensions, with details summarized in Table \ref{tab:downsampling}. 
We observe that increasing parameter counts via dimensional projection did not yield significant improvements in overall model performance. 
Consequently, we opt for the average pooling strategy for feature downsampling.
\begin{table}[!htbp]
    \caption{Ablation on different strategy for geometry token downsampling}
    \label{tab:downsampling}
    \centering
    \footnotesize
    \setlength{\tabcolsep}{6pt}
    \belowrulesep=0pt
    \aboverulesep=0pt
    \resizebox{0.9\linewidth}{!}{
        \renewcommand{\arraystretch}{1.2}
        \begin{tabular}{c|c|c|c}
            \toprule
            \textbf{Strategy} & \textbf{PAM} & \textbf{BHB} & \textbf{DBPE} \\
            \midrule
            mlp & \textbf{82.0±1.6} & 66.3±1.7 & 88.7±1.2 \\
            pool & 80.0±0.8 & \textbf{85.0±1.4} & \textbf{88.3±0.9} \\
            \midrule
            \textbf{Strategy} & \textbf{PAC} & \textbf{BSE} & \cellcolor{yellow!15} \textbf{AVG. ($\uparrow$)} \\
            \midrule
            mlp & \textbf{99.3±0.9} & \textbf{62.3±3.3} & \cellcolor{yellow!15} 79.7 \\
            pool & 94.3±2.3 & 52.3±2.5 & \cellcolor{yellow!15} \textbf{80.0} \\
            \bottomrule
        \end{tabular}
    }
    \vspace{-3mm}
\end{table}

\begin{figure*}[!htbp]
    \centering
    \begin{subfigure}[b]{0.19\textwidth}
      \includegraphics[width=\textwidth]{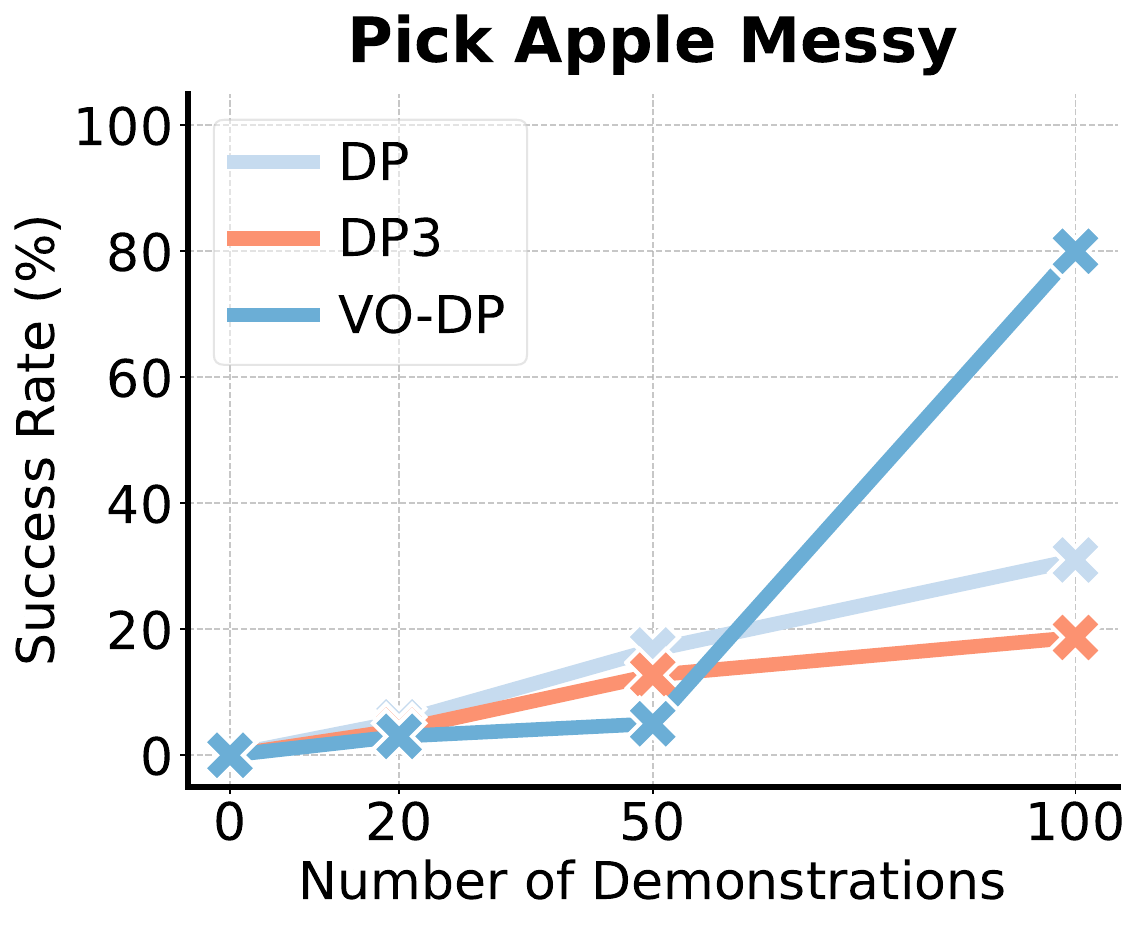}
      \caption{}
      \label{fig:PAM}
    \end{subfigure}%
    ~
    \begin{subfigure}[b]{0.19\textwidth}
      \includegraphics[width=\textwidth]{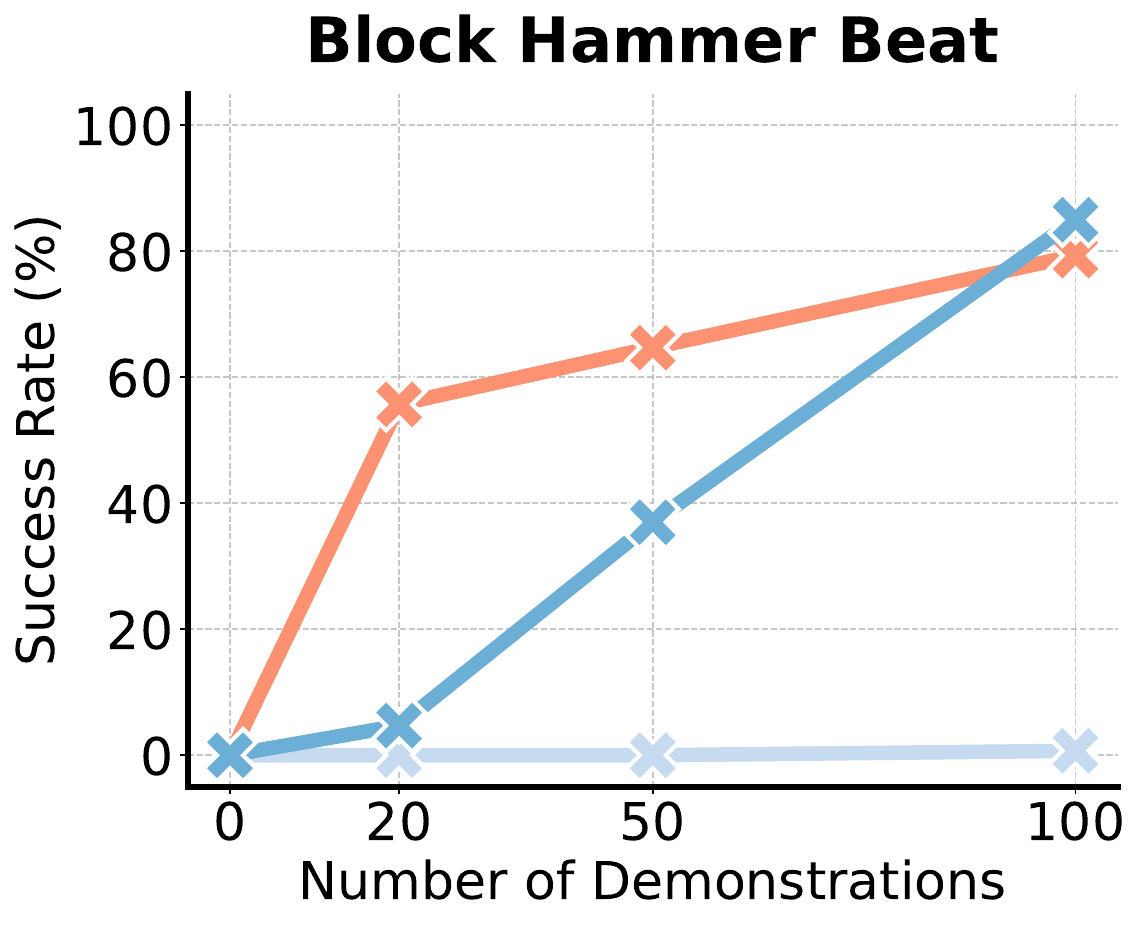}
      \caption{}
      \label{fig:BHB}
    \end{subfigure}
    ~
    \begin{subfigure}[b]{0.19\textwidth}
      \includegraphics[width=\textwidth]{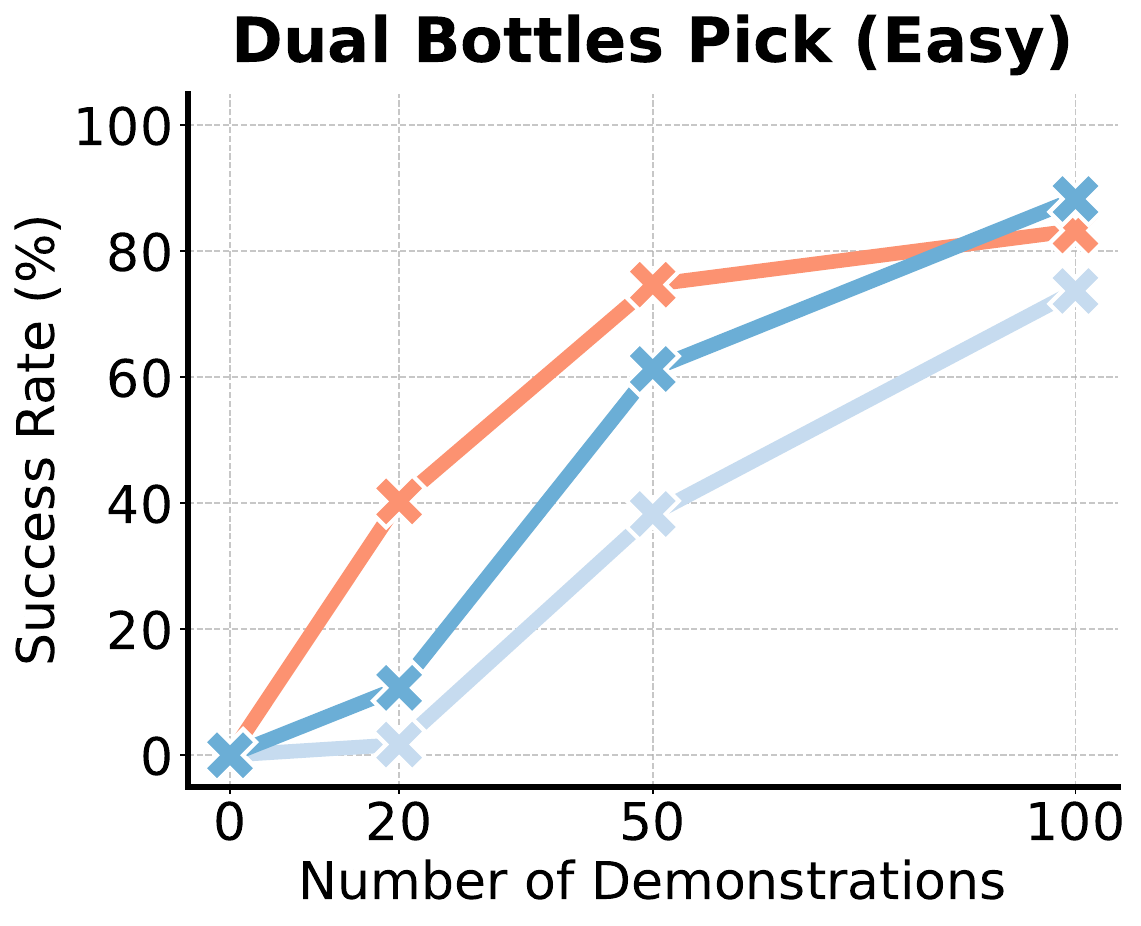}
      \caption{}
      \label{fig:DBPE}
    \end{subfigure}
    \begin{subfigure}[b]{0.19\textwidth}
      \includegraphics[width=\textwidth]{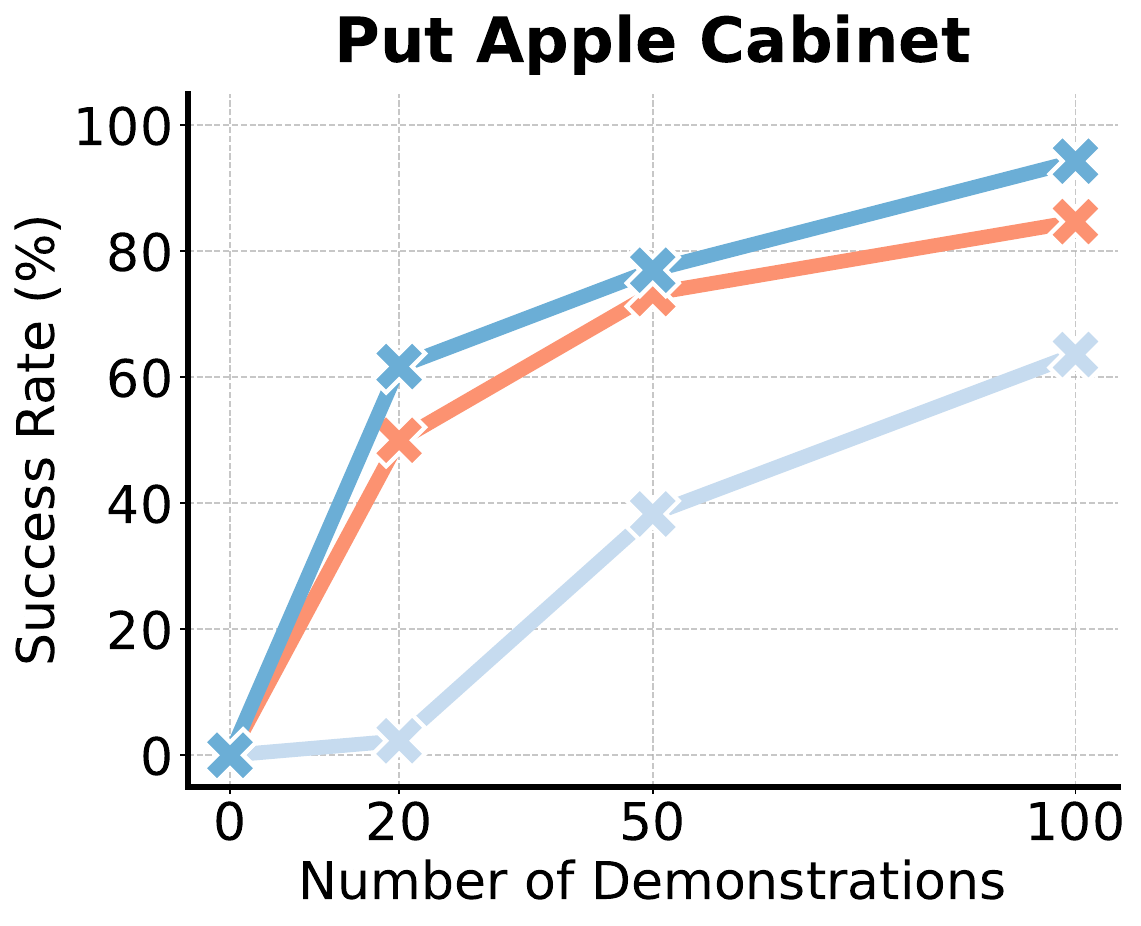}
      \caption{}
      \label{fig:PAC}
    \end{subfigure}
    \begin{subfigure}[b]{0.19\textwidth}
      \includegraphics[width=\textwidth]{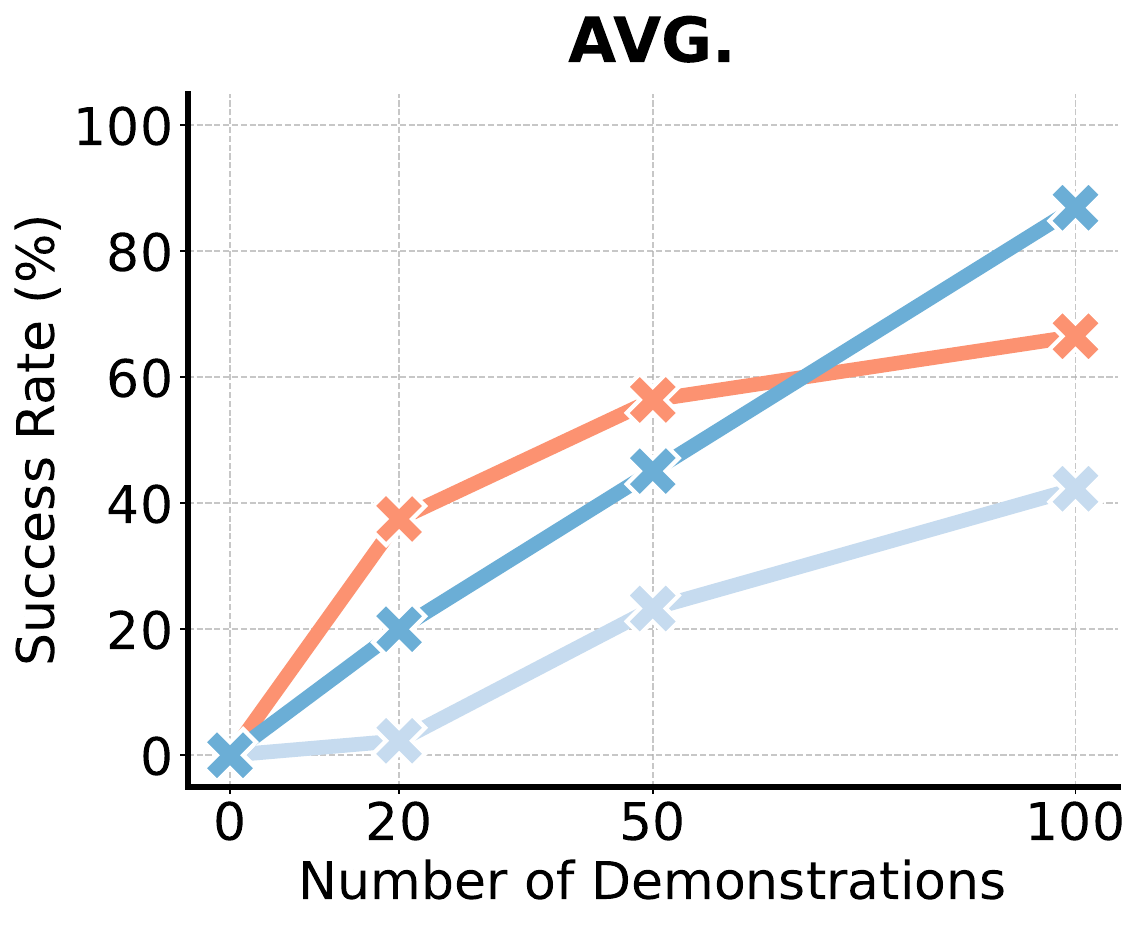}
      \caption{}
      \label{fig:AVG}
    \end{subfigure}
    \caption{Data efficiency and scaling capability comparison of VO-DP with baseline methods (DP and DP3) across four tasks. The table presents success rate changes as training demonstrations scale from 20 to 100, highlighting VO-DP’s more substantial performance improvements.}
    \label{fig:scaling}
    \vspace{-5mm}
\end{figure*}
\textbf{Efficient scaling with demonstrations.} 
As shown in Fig.~\ref{fig:scaling}, VO-DP exhibits high data efficiency and strong scaling capability. Compared with baselines (DP and DP3), it delivers more substantial performance gains as training demonstrations increase from 20 to 100, especially in high-complexity scenarios.
For example, in \textit{Pick Apple Messy}, its success rate jumps from 3.0\% to 80.0\%, outperforming DP and DP3 markedly.
A similar trend appears in \textit{Block Hammer Beat}: VO-DP’s success rate rises from 4.7\% to 85.0\%, while DP3 only improves modestly and DP gains little from more data. These results confirm that the strong prior knowledge in its pretrained visual encoder enables VO-DP to learn and generalize more effectively with limited demonstration data.

\begin{figure*}[!htbp]
    \centering
    \includegraphics[width=\textwidth]{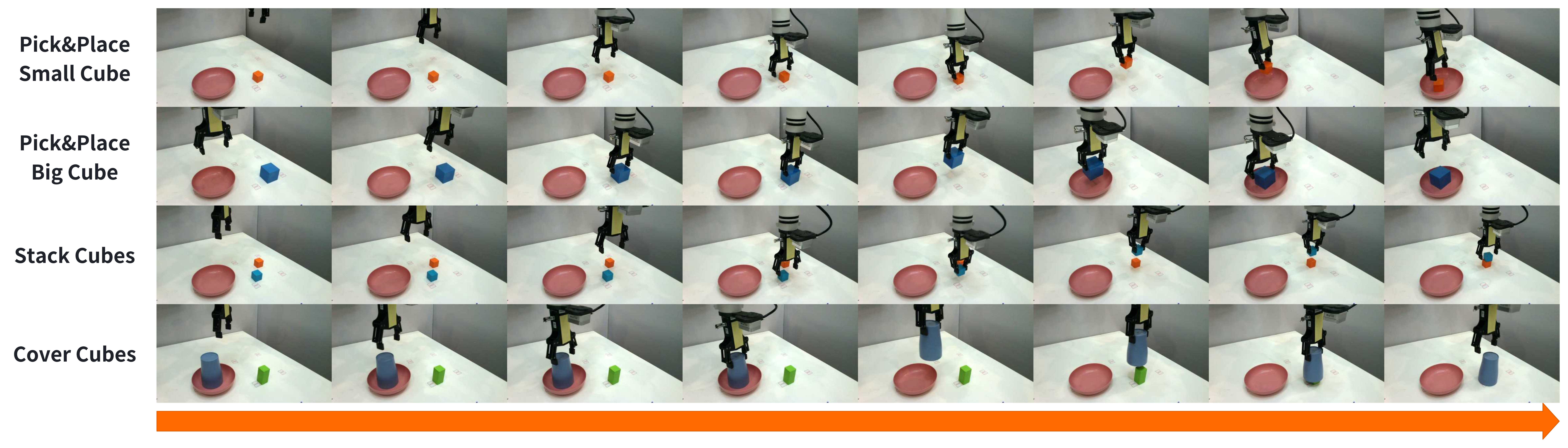}
    \vspace{-5mm}
    \caption{Visualize of 4 real-world tasks: Pick\&Place Small Cube (PPSC), Pick\&Place Big Cube (PPBC), Cover Cuboid (CC), Stack Cubes (SC).}
    \vspace{-5mm}
    \label{fig:realworld_vis}
\end{figure*}
\section{REAL WORLD EXPERIMENTS}
\subsection{Experiment Setup}

\begin{figure}[!htbp]
    \centering
    \includegraphics[width=\linewidth]{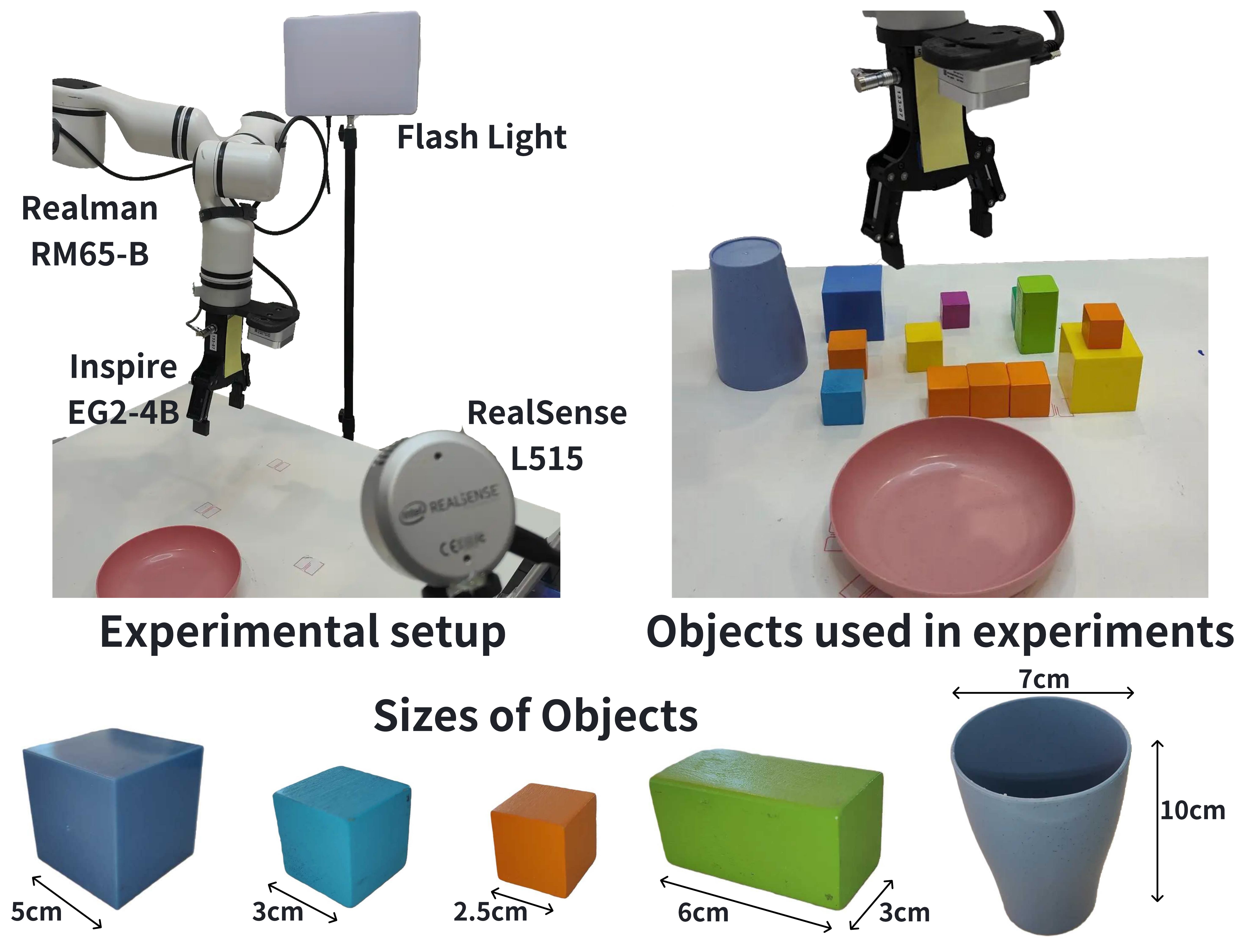}
    \caption{Information on the equipment used in real-world experiments and the objects involved in tasks.}
    \label{fig:setup}
\end{figure}

\textbf{Real robot benchmark.} 
We evaluate VO-DP on four real-world tasks and four robustness tests. 
As illustrated in Fig. \ref{fig:setup}, we use a Realman RM65-B robot equipped with an Inspire EG2-4C2 gripper, one RealSense L515 camera to capture real-world visual observations (containing both RGB images and point clouds) with robot states, and a controllable flashlight with adjustable color and frequency as an environmental disturbance source. 
All objects used in the experiments are shown in Fig. \ref{fig:setup}, which include multiple blocks and containers of varying shapes, sizes, and colors.
We now briefly describe the four spatial tasks:
\begin{itemize}
    \item \textbf{Pick\&Place Small Cube (PPSC).} Grasp a 3 cm cube and place it at the center of the plate.
    \item \textbf{Pick\&Place Big Cube (PPBC).} Grasp a 5 cm cube and place it at the center of the plate.
    \item \textbf{Cover Cuboid (CC).} Pick up a cup from the plate and move it to cover an upright 3cm×3cm×6cm cuboid.
    \item \textbf{Stack Cubes (SC).} Stack a blue 3 cm cube on top of an orange 3 cm cube.
\end{itemize}
All the tasks are visualized in Fig.~\ref{fig:realworld_vis}.

\begin{figure*}[!htbp]
    \centering
    \includegraphics[width=0.9\textwidth]{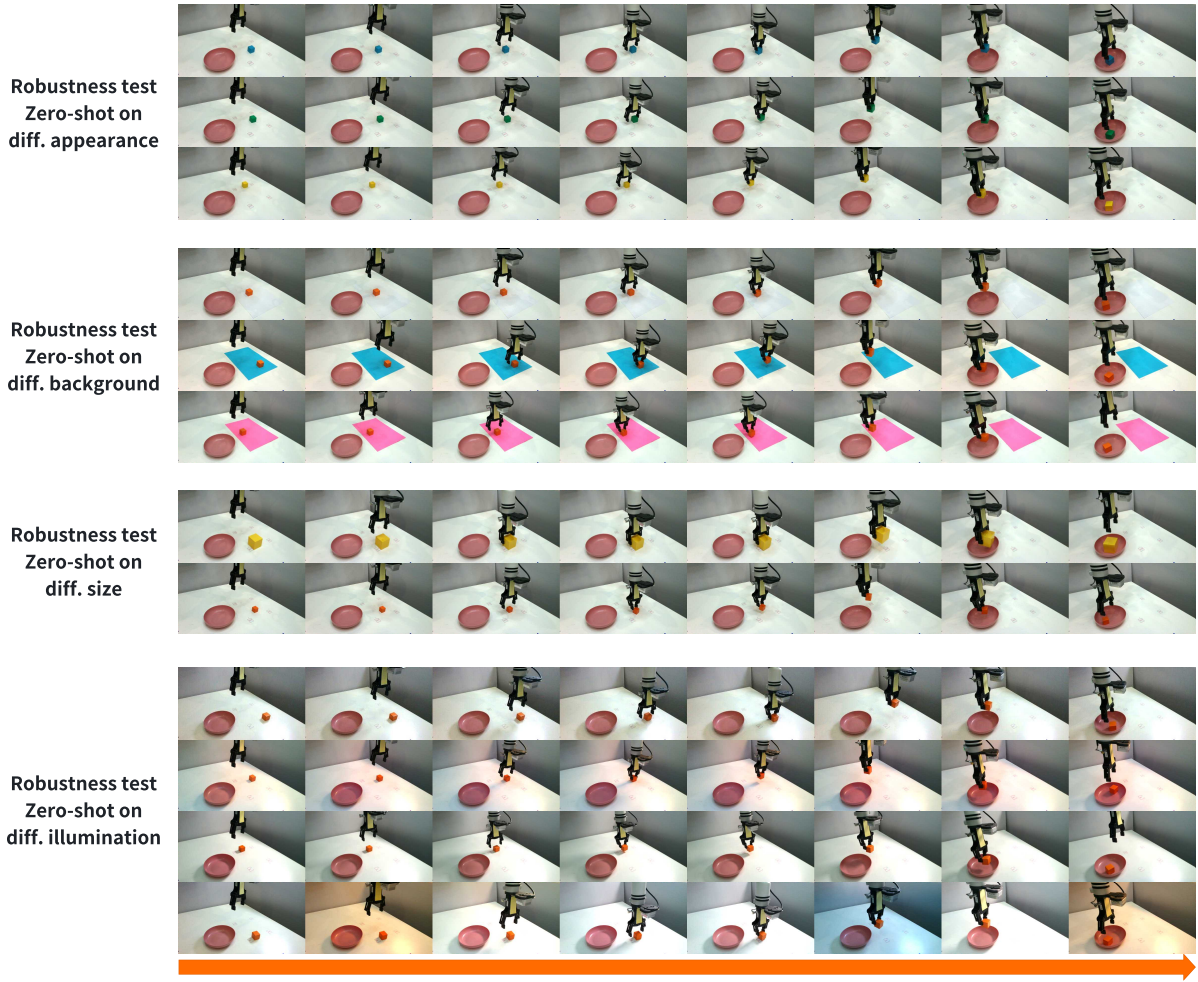}
    \caption{Visualization of the 4 real-world robustness tests: Zero-Shot Testing on Cube Appearance, Cube Size, Background Color, and Illumination.}
    \vspace{-5mm}
    \label{fig:robust_vis}
\end{figure*}

Additionally, we design four robustness tests based on the \textit{Pick\&Place Small cube} task: 
\begin{itemize}
    \item \textbf{Size robustness.} Train: 3 cm cube; Test: cubes of 2.5 cm, 3 cm and 5 cm. 
    \item \textbf{Appearance robustness.} Train: orange cubes; Test: cubes of all colors. 
    \item \textbf{Illumination robustness.} Train: normal ambient lighting; Test: normal ambient lighting and strobe lighting. 
    \item \textbf{Background robustness.} Train: a standard desktop surface; Test: a standard desktop surface and ones covered with colored paper. 
\end{itemize}
All the tests are visualized in Fig.~\ref{fig:robust_vis}


\begin{figure}[!thbp]
    \centering
    \includegraphics[width=0.7\linewidth]{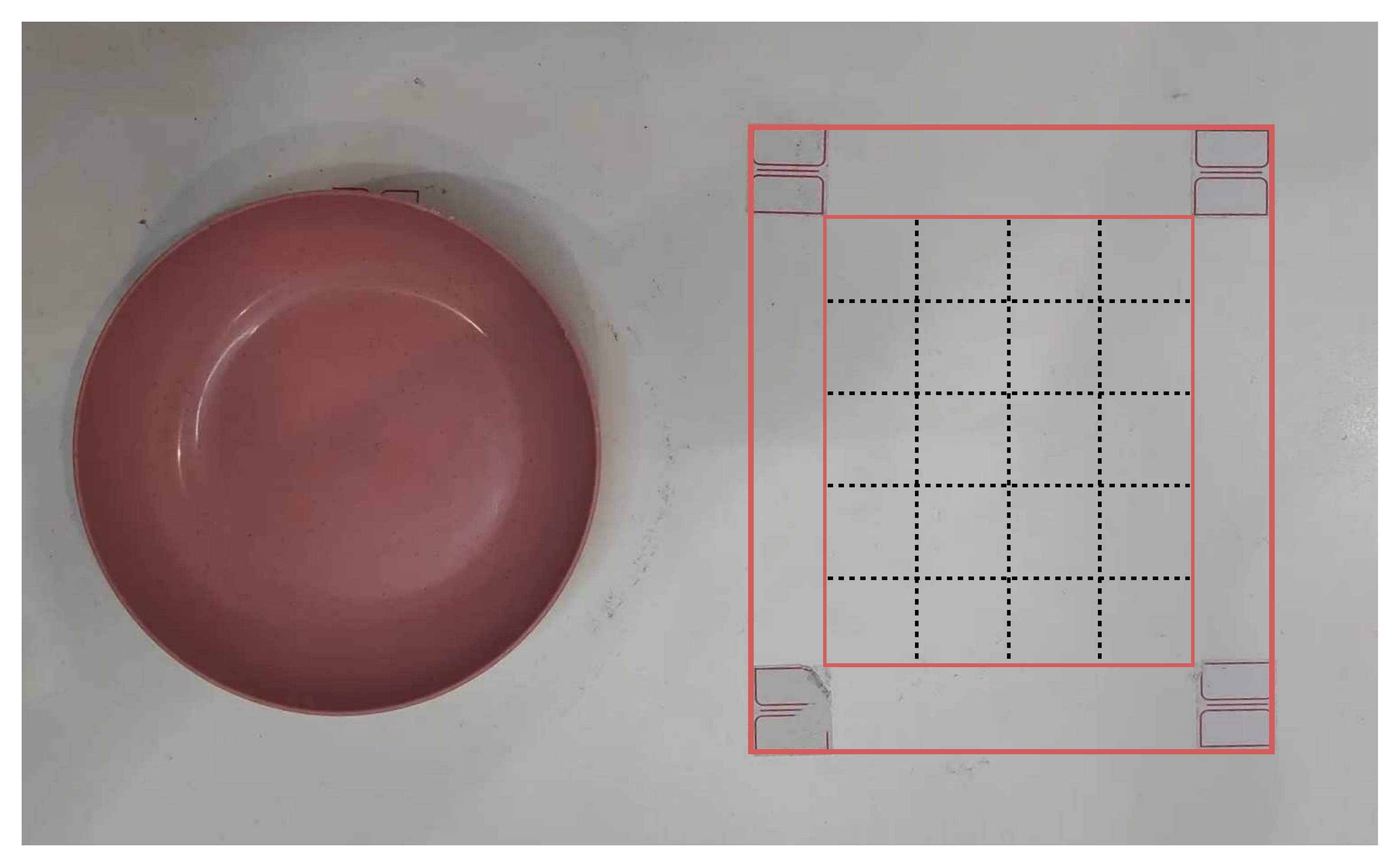}
    \caption{Desktop layout: a plate is placed on the left side, and an object placement area is on the right side. The placement area is divided into a 4×4 grid, with each grid cell measuring approximately 3 cm.}
    \vspace{-3mm}
    \label{fig:layout}
\end{figure}

\textbf{Data Collection.} We collect 200 demonstrations per task using the teleoperation device provided with the Realman robot. 
The operational area is uniformly partitioned into multiple grids, as illustrated in Fig. \ref{fig:layout}. 
During data collection, the target object is sequentially placed at random positions within each grid. For instance, in the \textit{Stack cubes} task, the orange cube is placed in distinct grids one after another, while the blue cube is positioned in all remaining feasible grids, thus ensuring coverage of all possible combinations.
During testing, the same strategy is employed to ensure a uniform spatial distribution for evaluation.

\textbf{Training Details.} 
As in the simulation experiments, we select DP and DP3 as our baselines. 
Given the color sensitivity of the \textit{Stack Cubes} task, we use the color-variant version of DP3 for comparison. 
Based on the simulation results, we choose the single-frame VO-DP-1 for real-world evaluation (unless otherwise specified below, VO-DP refers to VO-DP-1 by default). 
It is noteworthy that the point clouds processing in DP3 depends on a manually defined operational region—an approach that proves impractical for real-world robotic manipulation scenarios.

\begin{table}[!htbp]
    \caption{Real-world Performance. In the four real-world tasks, it can be observed that VO-DP significantly outperforms the other two methods.}
    \label{tab:realworld}
    \centering
    \footnotesize
    \setlength{\tabcolsep}{6pt}
    \belowrulesep=0pt
    \aboverulesep=0pt
    \resizebox{1.0\linewidth}{!}{
        \renewcommand{\arraystretch}{1.2}
        \begin{tabular}{c|c|c|c|c| >{\columncolor{yellow!15}} c}
            \toprule
            \textbf{Method} & \textbf{PPSC} & \textbf{PPBC} & \textbf{CC} & \textbf{SC} & \textbf{AVG. ($\uparrow$)} \\
            \midrule
            DP      & 23.3 & 16.7 & 3.3  & 1.7  & 11.2±9.1 \\
            \rowcolor{red!15} DP3     & 73.3 & 68.3 & 75.0 & 53.3 & 67.5±8.5 \\
            \rowcolor{cyan!15} VO-DP-1 & \textbf{96.7} & \textbf{91.7} & \textbf{93.3} &\textbf{70.0} & \textbf{87.9±10.5} \\
            \bottomrule
        \end{tabular}
    }
    \vspace{-3mm}
\end{table}
\subsection{Real-world Performance}
Results for our real robot tasks are given in Table \ref{tab:realworld}. 
Our experimental results demonstrate that VO-DP achieves strong performance and generalization capability in real-world physical environments. 
It attains an average success rate of 87.9\% across all four tasks, significantly outperforming the point cloud-based method DP3 (67.5\%) and the conventional vision-only approach DP (11.2\%).
Specifically, VO-DP achieves the highest performance in \textit{Cover cuboid} and \textit{Pick\&Place cube}. 
This demonstrates that the visual encoder of VO-DP possesses the capability to extract robust, task-discriminative features—thereby facilitating accurate perception of object geometry, spatial relationships, and manipulation intent.
The results confirm that VO-DP achieves effective transfer from simulation to complex real-world environments.
Notably, it outperforms the method DP3 dependent on expensive depth sensors, while only utilizing a low-cost RGB camera.
We attribute the degradation of DP3’s real-world performance, relative to its performance in simulations, to the lack of idealized sensing conditions. 
In real-world scenarios, depth sensors are inherently affected by noise, calibration inaccuracies, viewpoint dependence, and artifacts introduced during point clouds preprocessing. 
All these factors collectively lead to the performance gap of DP3 between simulation and real-world scenarios.

\subsection{Robustness Test}
\textbf{Size Robustness.} 
To evaluate the geometric robustness of VO-DP, we test the model trained on $3.0\, cm$ cubes using $2.5\, cm$, $3.0\, cm$ and $5.0\, cm$ cubes, as illustrated in Fig. \ref{fig:setup}. 
The evaluation involves randomized tests across 20 grids in the central operational area, with results summarized in Table \ref{tab:size-robust}. 
Experimental results show that VO-DP generalizes geometrically to unseen object sizes, maintaining robust performance across both smaller and larger objects with an average success rate of 65.0\%. 
This indicates that the visual encoder captures geometric and spatial representations that exhibit scale-invariant generalization.
\begin{table}[!htbp]
    \caption{Size Robustness. Train: 3.0 cm cubes; Test: 2.5 cm, 3.0 cm and 5 cm cubes.}
    \label{tab:size-robust}
    \centering
    \footnotesize
    \setlength{\tabcolsep}{6pt}
    \belowrulesep=0pt
    \aboverulesep=0pt
    \resizebox{0.8\linewidth}{!}{
        \renewcommand{\arraystretch}{1.2}
        \begin{tabular}{c|c|c| >{\columncolor{yellow!15}} c}
            \toprule
            \textbf{3.0 cm} & \textbf{2.5 cm} & \textbf{5.0 cm} & \textbf{AVG.} \\
            \midrule
            85.0 & 60.0 & 50.0 & \textbf{65.0±14.7}\\
            \bottomrule
        \end{tabular}
    }
    \vspace{-3mm}
\end{table}

\textbf{Appearance Robustness.} 
We evaluate the robustness of VO-DP to varying object appearances by testing the model trained on orange cubes using blue, green, and yellow cubes. 
The results are presented in Table \ref{tab:color-robust}.
VO-DP achieves strong performance on the yellow cube, which is chromatically close to the training color demonstrating its capacity for color generalization.
However, performance declines markedly on distantly colored objects such as blue and green, suggesting that semantic color understanding remains partially constrained by the training distribution.
\begin{table}[!htbp]
    \caption{Appearance Robustness. Train: \textcolor{orange!80}{\rule{5pt}{5pt}} cubes; Test: \textcolor{orange!80}{\rule{5pt}{5pt}}, \textcolor{cyan!70}{\rule{5pt}{5pt}}, \textcolor{green!70}{\rule{5pt}{5pt}}, \textcolor{yellow}{\rule{5pt}{5pt}} cubes.}
    \label{tab:color-robust}
    \centering
    \footnotesize
    \setlength{\tabcolsep}{6pt}
    \belowrulesep=0pt
    \aboverulesep=0pt
    \resizebox{0.8\linewidth}{!}{
        \renewcommand{\arraystretch}{1.2}
        \begin{tabular}{c|c|c|c| >{\columncolor{yellow!15}} c}
            \toprule
            \textcolor{orange!80}{\rule{5pt}{5pt}} & 
            \textcolor{cyan!70}{\rule{5pt}{5pt}} & 
            \textcolor{green!70}{\rule{5pt}{5pt}} & 
            \textcolor{yellow}{\rule{5pt}{5pt}} & 
            \textbf{AVG.} \\
            \midrule
            85.0 & 50.0 & 40.0 & 90.0 & \textbf{66.3±21.6}\\
            \bottomrule
        \end{tabular}
    }
    \vspace{-3mm}
\end{table}

\begin{figure}[!htbp]
    \centering
    \includegraphics[width=0.9\linewidth]{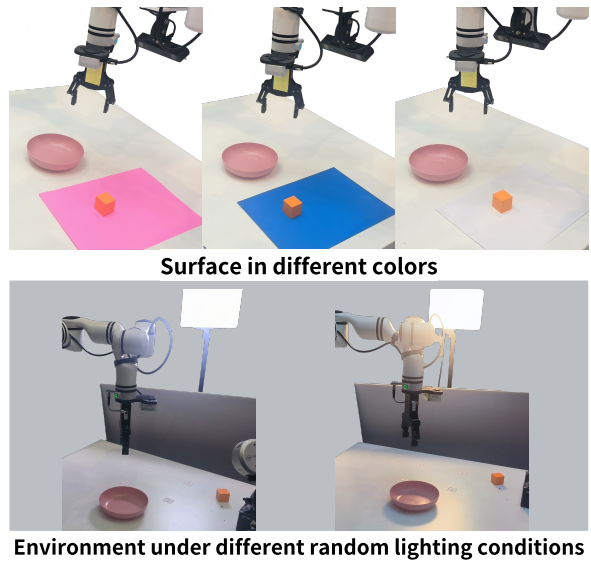}
    \caption{Different environments were utilized for the robustness tests. We covered the original table surface with pink, blue, and white paper, respectively. We also employed lighting with random color temperatures and brightness to alter the ambient illumination.}
    \vspace{-5mm}
    \label{fig:environment}
\end{figure}

\textbf{Illumination Robustness.} 
We validate the robustness of VO-DP to varying lighting conditions by adjusting flashlight settings, as shown in \ref{fig:environment}.
In the \textit{Light Switch} test, both intensity and color temperature are randomly configured for each evaluation position.
During the \textit{Blinking} test, a low-frequency blinking mode continuously alters the ambient illumination.
Results are presented in Table \ref{tab:light-robust}.
The results indicate that VO-DP maintains strong robustness under challenging illumination variations.
Under extreme conditions such as stochastic light switching and low-frequency blinking, it achieves performance comparable to that under standard lighting, with an average success rate of 83.3\%.
This confirms the ability of the method to extract illumination-invariant visual representations and its notable resilience to variations in color, brightness, and dynamic lighting interference.
\begin{table}[!htbp]
    \caption{Illumination Robustness. Train: Normal; Test:Nomal, Light Switch, Blinking.}
    \label{tab:light-robust}
    \centering
    \footnotesize
    \setlength{\tabcolsep}{6pt}
    \belowrulesep=0pt
    \aboverulesep=0pt
    \resizebox{0.9\linewidth}{!}{
        \renewcommand{\arraystretch}{1.2}
        \begin{tabular}{c|c|c| >{\columncolor{yellow!15}} c}
            \toprule
            \textbf{Normal} & \textbf{Light Switch} & \textbf{Blinking} & \textbf{AVG.} \\
            \midrule
            85.0 & 80.0 & 85.0 & \textbf{83.3±2.4}\\
            \bottomrule
        \end{tabular}
    }
    \vspace{-3mm}
\end{table}

\textbf{Background Robustness.}
Background generalization presents a significantly greater challenge for methods relying on RGB image inputs. 
To evaluate this, we cover the operational area with white, pink, and blue paper respectively during testing, as shown in \ref{fig:environment}. 
The results are shown in Table \ref{tab:bg-robust}.
The results indicate that VO-DP generalizes effectively across substantial variations in background appearance.
The model achieves high manipulation success rates under diverse background colors demonstrating robust performance despite visual domain shifts.
\begin{table}[!htbp]
    \caption{Background Robustness. Train: desktop surface; Test: desktop surface, \textcolor{lightgray}{\rule{5pt}{5pt}}, \textcolor{pink}{\rule{5pt}{5pt}} and \textcolor{blue}{\rule{5pt}{5pt}} surface. }
    \label{tab:bg-robust}
    \centering
    \footnotesize
    \setlength{\tabcolsep}{6pt}
    \belowrulesep=0pt
    \aboverulesep=0pt
    \resizebox{0.9\linewidth}{!}{
        \renewcommand{\arraystretch}{1.2}
        \begin{tabular}{c|c|c|c| >{\columncolor{yellow!15}} c}
            \toprule
            \textbf{desktop surface} & 
            \textcolor{lightgray}{\rule{5pt}{5pt}} & 
            \textcolor{pink}{\rule{5pt}{5pt}} & 
            \textcolor{blue}{\rule{5pt}{5pt}} & 
            \textbf{AVG.} \\
            \midrule
            85.0 & 90.0 & 80.0 & 95.0 & \textbf{87.5±5.6} \\
            \bottomrule
        \end{tabular}
    }
    \vspace{-5mm}
\end{table}

\section{CONCLUSION}
This paper addresses the underexplored potential of vision-only approaches in visuomotor diffusion policy learning for robotic manipulation, proposing a single-view, vision-only method (VO-DP) that bridges the performance gap between vision-only and point cloud-based baselines.
VO-DP leverages pretrained visual foundation models to fuse semantic and geometric features effectively via targeted extraction, cross-attention fusion, and CNN compression, supporting downstream policy learning. Extensive experiments validate its efficacy: on the RoboTwin benchmark, it significantly outperforms the vision-only baseline DP and matches point cloud-based DP3, while achieving the highest average success rate in real-world tasks with strong robustness across conditions.
Overall, VO-DP demonstrates that vision-only approaches can achieve high accuracy and robustness in robotic manipulation without expensive depth sensors, highlighting their potential for cost-effective, scalable real-world deployment. Future work may extend VO-DP to multi-view settings or more complex dynamic manipulation tasks.

\textbf{Limitations.} While the proposed method has been validated for vision-only robotic manipulation, several limitations present opportunities for further investigation.
The use of a generic, pre-trained VGGT backbone may limit reconstruction accuracy in embodied scenarios.
The sparse feature distribution from the VGGT encoder increases the difficulty of policy learning. This issue is further exacerbated in multi-view reconstruction tasks.
The inference speed of VGGT is relatively slow, which can constrain the real-time responsiveness of a robot in practical deployments.





\bibliographystyle{ieeetr}
\bibliography{reference}

\clearpage
\newpage
\onecolumn
\begin{appendix}

\subsection{Real-World Detailed Results}
\begin{figure*}[!htbp]
    \centering
    \begin{subfigure}[b]{0.329\linewidth}
      \includegraphics[width=\textwidth]{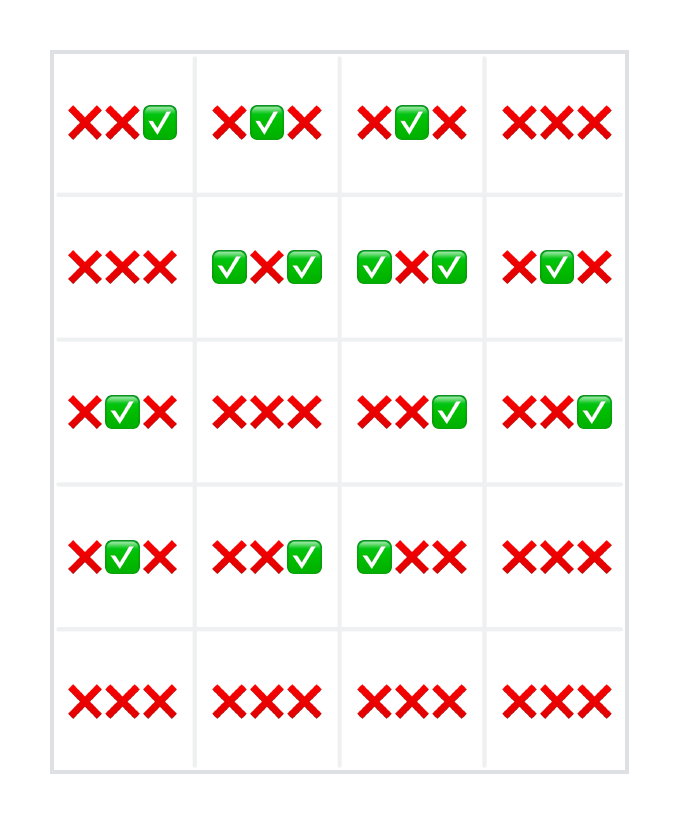}
      \caption{DP}
      \label{fig:PPSC-DP}
    \end{subfigure}%
    ~
    \begin{subfigure}[b]{0.329\linewidth}
      \includegraphics[width=\textwidth]{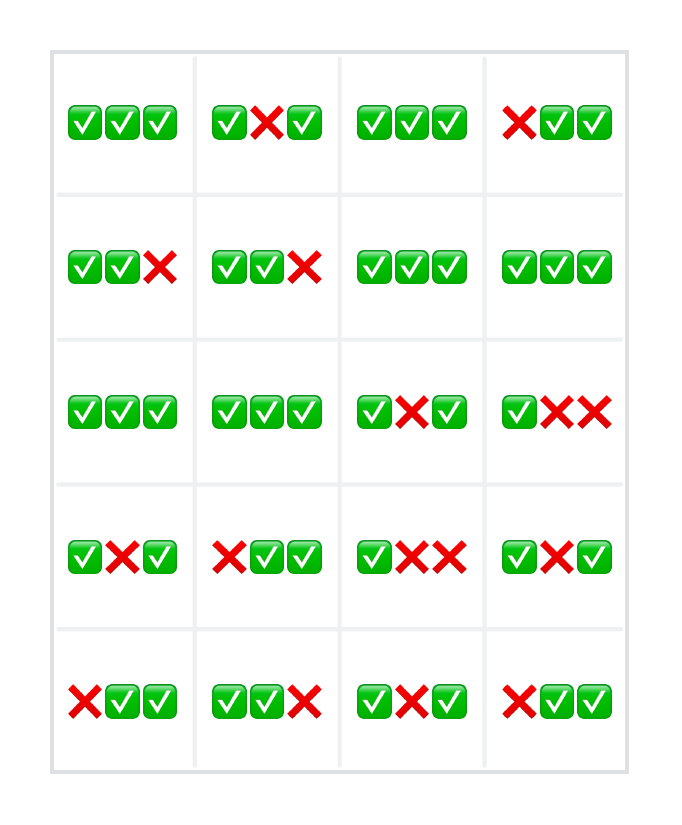}
      \caption{DP3}
      \label{fig:PPSC-DP3}
    \end{subfigure}
    \begin{subfigure}[b]{0.329\linewidth}
      \includegraphics[width=\textwidth]{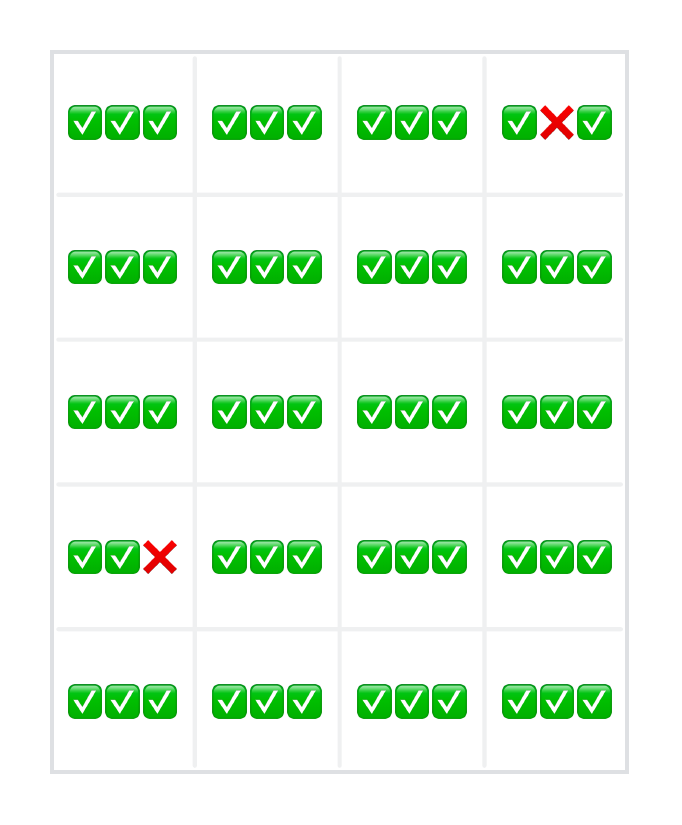}
      \caption{VO-DP-1}
      \label{fig:PPSC-VO-DP}
    \end{subfigure}
    \caption{The Pick\&Place Small Cube real-world task is uniformly distributed in the experimental space, with detailed results from three independent repetitions for each grid cell. \checkmark ~ indicates task success, while \texttimes ~ represents task failure. Among them, (a) DP achieved 14/60 successes (23.3\%); (b) DP3 succeeded in 44/60 trials (73.3\%); and (c) our method, VO-DP-1, attained the highest performance with 58/60 successes (96.7\%). }
    \label{fig:PPSC}
    \vspace{-5mm}
\end{figure*}

\begin{figure*}[!htbp]
    \centering
    \begin{subfigure}[b]{0.329\linewidth}
      \includegraphics[width=\textwidth]{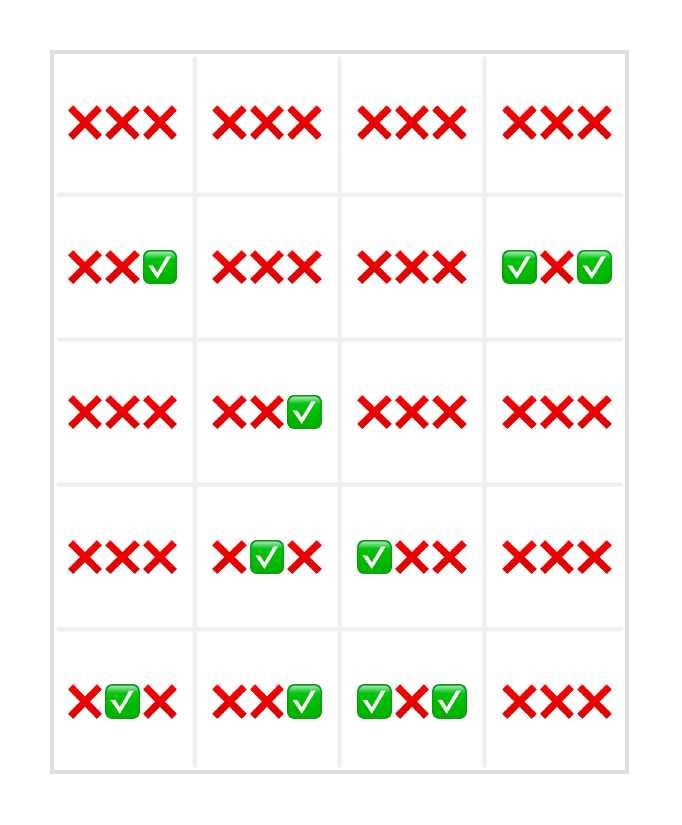}
      \caption{DP}
      \label{fig:PPBC-DP}
    \end{subfigure}%
    ~
    \begin{subfigure}[b]{0.329\linewidth}
      \includegraphics[width=\textwidth]{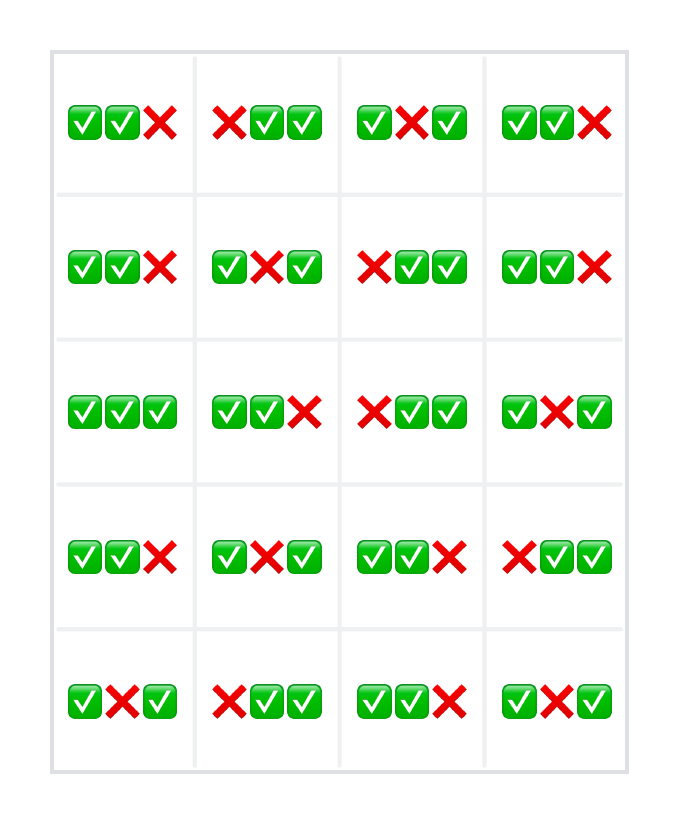}
      \caption{DP3}
      \label{fig:PPBC-DP3}
    \end{subfigure}
    \begin{subfigure}[b]{0.329\linewidth}
      \includegraphics[width=\textwidth]{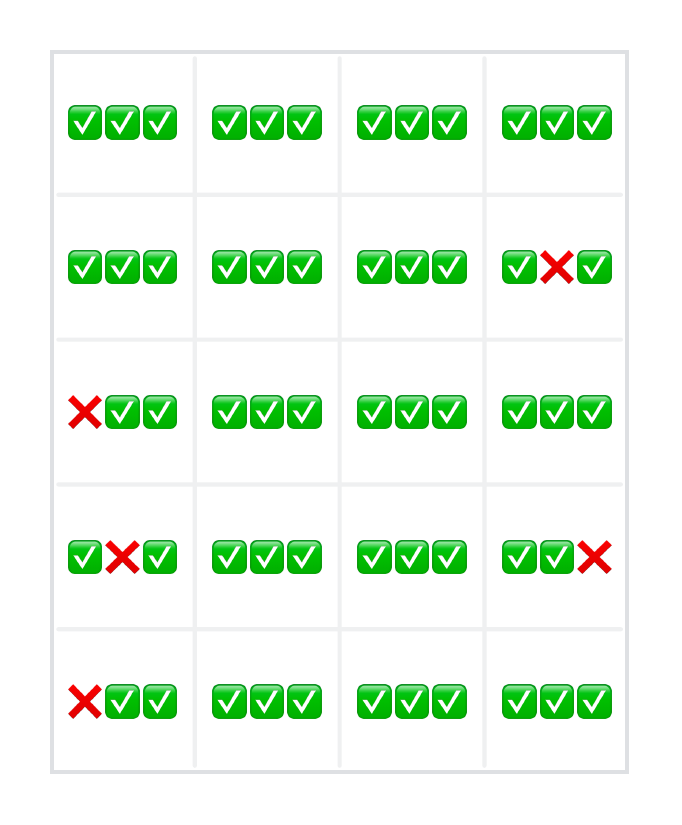}
      \caption{VO-DP-1}
      \label{fig:PPBC-VO-DP}
    \end{subfigure}
    \caption{The Pick\&Place Big Cube real-world task is uniformly distributed in the experimental space, with detailed results from three independent repetitions for each grid cell. \checkmark ~ indicates task success, while \texttimes ~ represents task failure. Among them, (a) DP achieved 10/60 successes (16.7\%); (b) DP3 succeeded in 41/60 trials (68.3\%); and (c) our method, VO-DP-1, attained the highest performance with 55/60 successes (91.7\%).}
    \label{fig:PPBC}
    \vspace{-5mm}
\end{figure*}

\begin{figure*}[!htbp]
    \centering
    \begin{subfigure}[b]{0.329\linewidth}
      \includegraphics[width=\textwidth]{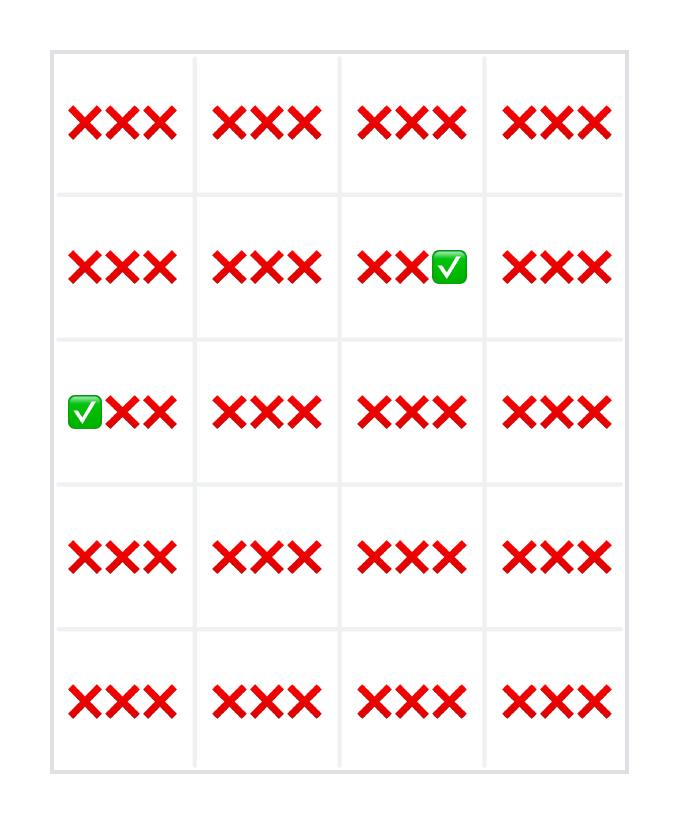}
      \caption{DP}
      \label{fig:CC-DP}
    \end{subfigure}%
    ~
    \begin{subfigure}[b]{0.329\linewidth}
      \includegraphics[width=\textwidth]{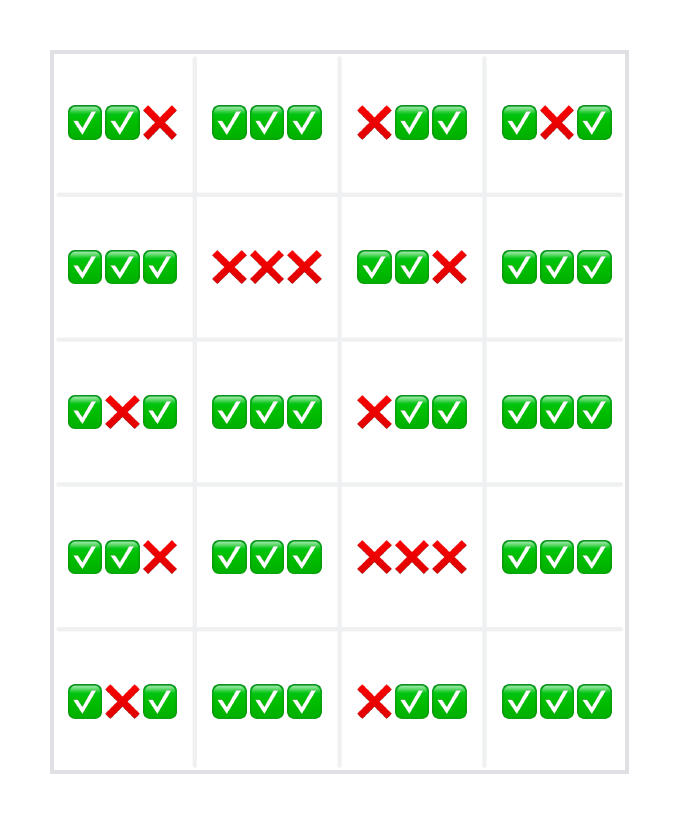}
      \caption{DP3}
      \label{fig:CC-DP3}
    \end{subfigure}
    \begin{subfigure}[b]{0.329\linewidth}
      \includegraphics[width=\textwidth]{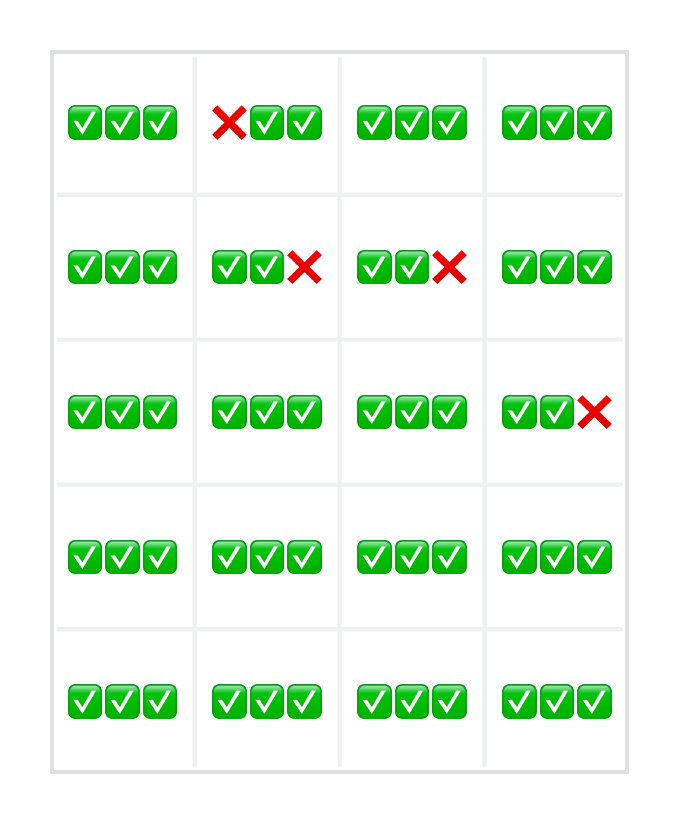}
      \caption{VO-DP-1}
      \label{fig:CC-VO-DP}
    \end{subfigure}
    \caption{The Cover Cube real-world task is uniformly distributed in the experimental space, with detailed results from three independent repetitions for each grid cell. \checkmark ~ indicates task success, while \texttimes ~ represents task failure. Among them, (a) DP achieved 2/60 successes (3.3\%); (b) DP3 succeeded in 45/60 trials (75.0\%); and (c) our method, VO-DP-1, attained the highest performance with 56/60 successes (93.3\%).}
    \label{fig:CC}
    \vspace{-5mm}
\end{figure*}

\begin{figure*}[!htbp]
    \centering
    \begin{subfigure}[b]{0.329\linewidth}
      \includegraphics[width=\textwidth]{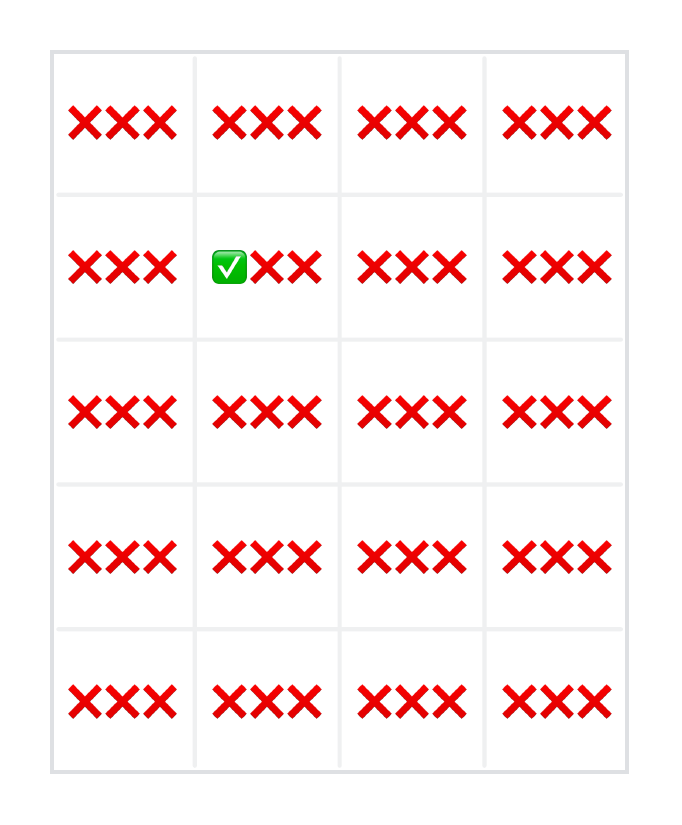}
      \caption{DP}
      \label{fig:SC-DP}
    \end{subfigure}%
    ~
    \begin{subfigure}[b]{0.329\linewidth}
      \includegraphics[width=\textwidth]{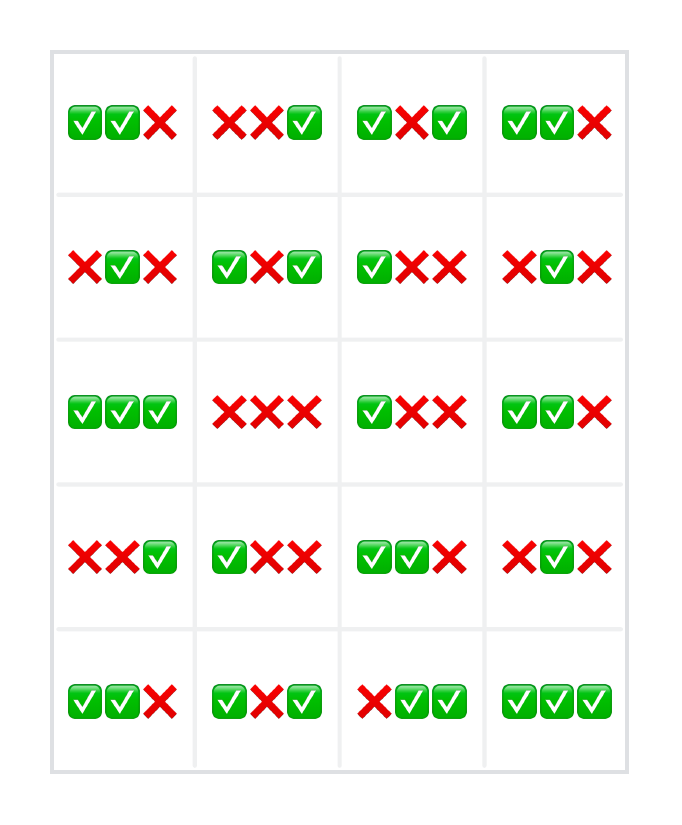}
      \caption{DP3}
      \label{fig:SC-DP3}
    \end{subfigure}
    \begin{subfigure}[b]{0.329\linewidth}
      \includegraphics[width=\textwidth]{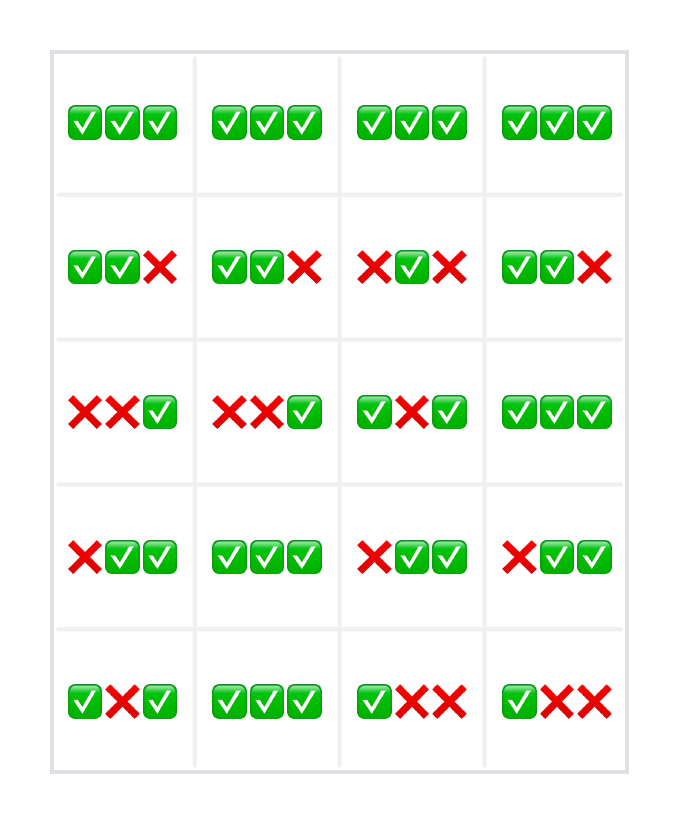}
      \caption{VO-DP-1}
      \label{fig:SC-VO-DP}
    \end{subfigure}
    \caption{The Stack Cubes real-world task is uniformly distributed in the experimental space, with detailed results from three independent repetitions for each grid cell. \checkmark ~ indicates task success, while \texttimes ~ represents task failure. Among them, (a) DP achieved 1/60 successes (3.3\%); (b) DP3 succeeded in 32/60 trials (53.3\%); and (c) our method, VO-DP-1, attained the highest performance with 42/60 successes (70.0\%).}
    \label{fig:SC}
    \vspace{-5mm}
\end{figure*}

\end{appendix}

\end{document}